\documentclass[11pt]{article}

\usepackage[final]{acl}

\usepackage{times}
\usepackage{latexsym}

\usepackage[T1]{fontenc}

\usepackage[utf8]{inputenc}

\usepackage{microtype}

\usepackage{inconsolata}

\usepackage{graphicx}
\usepackage{amsmath}
\usepackage{pifont}
\usepackage{amssymb}
\usepackage{bm}
\usepackage{booktabs}
\usepackage{multirow}
\usepackage{subcaption}
\usepackage{enumitem}

\usepackage{amsthm}
\usepackage{xurl}
\usepackage{tabularx}
\usepackage{ragged2e}
\usepackage{placeins}

\theoremstyle{definition}
\newtheorem{definition}{Definition}

\title{Rethinking Expert Trajectory Utilization in LLM Post-training for Mathematical Reasoning}
\author{
    Bowen Ding\textsuperscript{\rm 1,2}~,
    Yuhan Chen\textsuperscript{\rm 2~\thanks{\quad Work done at Westlake University as an intern.}}~,
    Jiayang Lyu\textsuperscript{\rm 2~\footnotemark[1]}~,
    Jiyao Yuan\textsuperscript{\rm 4}~,
    Qi Zhu\textsuperscript{\rm 4},
    Shuangshuang Tian\textsuperscript{\rm 2~\footnotemark[1]}, 
  \\  
  \bf Dantong Zhu\textsuperscript{\rm 2~\footnotemark[1]}~,
    Futing Wang\textsuperscript{\rm 1,2}~, 
    Heyuan Deng\textsuperscript{\rm 4}~,
    Fei Mi\textsuperscript{\rm 4}\footnotemark[2]~~, 
    Lifeng Shang\textsuperscript{\rm 4}~, 
    Tao Lin\textsuperscript{\rm 2,3~\thanks{\quad Corresponding authors.}}, 
  \\ [1mm]
    \textsuperscript{1} Zhejiang University~
    \textsuperscript{2} School of Engineering, Westlake University \\
    \textsuperscript{3} Institute of Advanced Technology, Westlake Institute for Advanced Study \\
    \textsuperscript{4} Huawei Noah's Ark Lab \\
    $^{2}$\texttt{\{dingbowen, wangfuting, lintao\}@westlake.edu.cn} \\
    $^{4}$\texttt{\{yuanjiyao1, zhuqi41, dengheyuan, mifei2, Shang.Lifeng\}@huawei.com} \\
}

\begin{document}
\maketitle
\begin{abstract}
Supervised Fine-Tuning (SFT) and Reinforcement Learning (RL) 
dominate the post-training landscape for mathematical reasoning, 
yet differ fundamentally in their reliance on expert trajectories. 
To understand the optimal way to harness 
these trajectories for maximizing performance, we propose the 
\textbf{Plasticity-Ceiling Framework}. This framework empirically grounds 
the post-training landscape by decomposing the final performance ceiling into 
the foundational SFT performance and the subsequent RL plasticity (i.e., the 
maximum improvement via RL). Through extensive benchmarking, we establish the 
\textbf{Sequential SFT-then-RL} pipeline as the superior standard, overcoming
the stability and premature convergence deficits inherent in synchronized approaches. 
Furthermore, we derive precise scaling guidelines: (1) Transitioning to RL at 
the \textbf{Stable or Mild Overfitting Regime} of SFT maximizes the final ceiling 
by securing a robust SFT foundation with substantial RL plasticity; (2) Refuting the 
``Less is More'' hypothesis in SFT-then-RL scaling, we demonstrate that \textbf{Data Scale} 
determines the primary post-training potential, while \textbf{Trajectory Difficulty} acts 
as a performance multiplier; and (3) The \textbf{Minimum Validation Loss of SFT} serves 
as a reliable indicator for selecting the expert trajectories that maximize the ultimate 
performance ceiling. Our findings provide actionable guidelines for extracting maximum 
value from expert trajectories. Code: ~\url{https://github.com/LINs-lab/RETU}.
\end{abstract}

\section{Introduction}
\begin{figure}[t]
    \centering
    \includegraphics[width=1\linewidth]{
        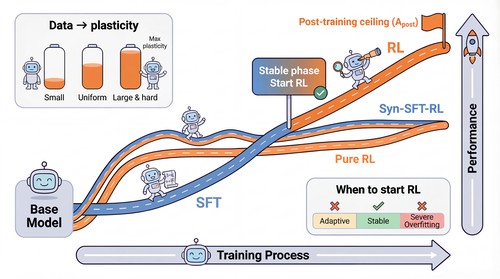
    }
    \caption{
\textbf{The post-training overview.}
    The sequential SFT-then-RL pipeline (blue$\rightarrow$orange) achieves the superior performance ceiling $A_{\text{post}}$, 
    outperforming both Pure-RL (orange) and Synchronized SFT-RL (striped) pathways. 
    Insets highlight that: (i) larger and more challenging datasets enhance RL plasticity (top-left); and 
    (ii) RL yields optimal results when initiated from the Stable Regime of SFT (bottom-right).}
    \label{fig:intro}
    \vspace{-0.5cm}
\end{figure}

Mathematical reasoning demands strict logic and complex planning, making it a critical benchmark for AI intelligence. 
Evolving generic Large Language Models (LLMs) into specialized Large Reasoning Models (LRMs) capable of tackling these tasks 
relies on effective post-training, often combining Supervised Fine-Tuning (SFT) and Reinforcement Learning (RL)~\citep{deepseek-math,qwen2_5_math}.

These paradigms diverge fundamentally in their reliance on expert trajectories. SFT fully relies on expert trajectories (prompt-solution pairs) to instill reasoning priors via 
imitation, while RL methods such as GRPO~\citep{deepseek-math} allow models to leverage prompt-answer pairs to self-explore reasoning paths through reward incentives, 
without the expert trajectory utilization. Despite the consensus on the necessity of both, a critical question remains unresolved: \textit{What is the optimal mechanism to utilize expert trajectories (i.e., SFT data) to 
maximize the post-training performance ceiling in mathematical reasoning?}

The methodology for effective expert trajectory utilization currently faces an unresolved paradigm dilemma. Recent works propose {Synchronized SFT-RL} (Syn-SFT-RL) algorithms, 
such as UPT \citep{lv2025unifiedviewlargelanguage}, SRFT \citep{Fu2025SRFT}, and LUFFY \citep{yan2025learningreasonoffpolicyguidance}, which integrate imitation loss directly 
into the RL optimization loop. While these methods often claim superior efficiency over sequential approaches, this advantage is critically constrained by their reliance on 
limited SFT data (only about 46K). This raises a fundamental question: whether Syn-SFT-RL maintains its claimed superiority when measured against the SFT and sequential SFT-then-RL baselines 
thoroughly optimized on significantly larger expert trajectory datasets.

Conversely, some LLM practitioners~\citep{qwen3, glm45, deepseek-r1} typically favor the straightforward sequential {SFT-then-RL} pipeline. However, the principles governing this 
successful approach remain largely empirical and lack systematic definition in two critical areas. First, concerning the \textbf{Optimal Timing} for switching from SFT to RL, the 
criteria lack a formal guidance. Second, regarding \textbf{Data Properties}, although the ``Less is More''~\citep{limo,s1} approach achieves comparable SFT accuracy with minimal 
data, it is unclear whether this compromises the subsequent RL scaling potential or leads to premature convergence. Similarly, while harder data push SFT boundaries~\citep{tong2024dartmath, d3}, 
its precise influence on the overall post-training ceiling remains unclarified. Consequently, these tensions highlight the urgent need for a unified framework to understand how the characteristics 
of SFT data dictate the entire post-training performance.

To rigorously address these systemic gaps, we propose a {Plasticity-Ceiling} analytical framework in \S~\ref{sec:framework}. This framework provides a unified view of all paradigms and enables the 
quantitative decomposition of the theoretical performance ceiling ($A_{\text{post}}$) into two measurable components: the\textbf{ SFT Performance} ($P_{\text{sft}}$) achieved under SFT compute $x_{\text{sft}}$, 
and the remaining \textbf{RL Plasticity} ($PL_{\text{rl}}$), which represents the maximum potential for subsequent RL improvement.

By conducting extensive experiments with the large-scale (i.e., 889K unique samples in maximum) SFT data on Qwen2.5-7B~\citep{qwen2025qwen25technicalreport} and validating on Llama3.2-3B~\citep{llama32} on the test set
including six popular mathematical benchmarks, we demystify expert trajectory utilization and establish a rigorous standard for post-training scaling in mathematical reasoning: 
\ding{202} \textbf{Sequential Paradigm Dominance} (\S~\ref{sec:paradigm_compare}).
 We empirically establish the superiority of the {Sequential SFT-then-RL} pipeline over the unstable, sensitive Syn-SFT-RL approaches, as 
well as pure SFT and RL. A robust SFT phase is necessary to establish the foundational SFT performance ($P_{\text{sft}}$) and unlock the maximum plasticity ($PL_{\text{rl}}$) of subsequent RL. 
\ding{203} \textbf{Switch RL until SFT Saturation} (\S~\ref{sec:compute_impact}). 
We identify the Stable or Mild-Overfitting Regime of validation loss saturation as the optimal SFT-to-RL transition window, 
both for the in-domain and the out-of-domain tasks.
\ding{204} \textbf{Scale and Difficulty Extend Ceiling} (\S~\ref{sec:data_impact}). 
We refute the ``Less is More'' hypothesis in the context of {SFT-then-RL} scaling. While minimal data yields SFT efficiency, the SFT data 
scale remains the primary determinant of the final ceiling, while the trajectory difficulty acts as a multiplier. Furthermore, the \textbf{Minimum 
Validation Loss of SFT} serves as a robust predictor of the final post-training ceiling.

Our contributions are summarized as follows: 
\ding{202} We propose the \textbf{{Plasticity-Ceiling Framework}}, a theoretical mechanism that decomposes post-training performance into 
realized SFT performance and the subsequent RL plasticity to guide paradigm selection. 
\ding{203} We systematically benchmark diverse training strategies, identifying the \textbf{Sequential SFT-then-RL} pipeline 
as the rigorous standard for stability and performance over other approaches. 
\ding{204} We formulate \textbf{actionable insights} in SFT-then-RL pipeline, linking data properties and training dynamics to the final 
reasoning ceiling to enable more effective post-training development for mathematical reasoning.

\section{Related Works}
\noindent \textbf{Post-Training Paradigms.}
Post-training primarily relies on SFT and RL. While theoretical works attempt to unify them~\citep{roadsleadlikelihoodvalue, sft_dpo_unify}, they exhibit distinct empirical behaviors regarding generalization and distribution shifts~\citep{understand_trans, rl_razor}.
The sequential SFT-then-RL strategy is the industrial standard~\citep{practicaltwostagerecipemathematical, Vattikonda2025WebAgent}, though optimizing the transition is non-trivial; \citet{quagmires_sft_rl} caution that high SFT scores can be misleading, as over-fitted models may fail to improve during RL.
Alternatively, {Synchronized SFT-RL} methods like LUFFY~\citep{yan2025learningreasonoffpolicyguidance}, UPT~\citep{lv2025unifiedviewlargelanguage}, SRFT~\citep{Fu2025SRFT} and Prefix-RFT~\citep{Huang2025PrefixRFT} integrate imitation directly into RL to boost efficiency. Our work systematically compares these paradigms to identify the optimal mechanism for maximizing the performance ceiling.

\noindent \textbf{Expert Trajectories Utilization.}
The properties of SFT data critically influence the post-training. 
Regarding \textbf{scale}, a ``Less is More'' philosophy suggests that minimal, high-quality data suffices for SFT~\citep{limo, s1}. However, others argue that scale remains essential for complex reasoning~\citep{climb_ladder}.
Regarding \textbf{difficulty}, methods like MetaMath~\citep{yu2023metamath} and D3~\citep{d3} demonstrate that harder, difficulty-aware data selection improves SFT outcomes.
Crucially, prior works often evaluate SFT in isolation. We extend this inquiry to the RL phase, investigating how SFT data scale and difficulty dictate the model's {plasticity} (its headroom for subsequent RL scaling) rather than just immediate accuracy after SFT.

\section{Preliminary}
\label{sec:preliminary}
In this section, we summarize algorithms of various post-training paradigms, each characterized by distinct approaches to 
utilizing expert trajectories. Details are in Appx.~\ref{appx:post_train_algos}.

\noindent \textbf{SFT} 
optimizes the policy $\pi_{\theta}$ via imitation 
learning on expert trajectory pairs $(\mathbf{q},\bm{\tau})$ from $\mathcal{D}_{\text{SFT}}$~\citep{LmrlGym}. 
While it reliably instills expert knowledge, its performance is strictly bounded by the quality of thetraining 
data and lacks self-exploration capabilities~\citep{NEURIPS2022_b1efde53}.

\noindent \textbf{RL (GRPO \& DAPO)} 
enhances the model's reasoning capability via reward-guided exploration. 
In contrast to SFT, standard RL algorithms such as GRPO~\citep{deepseek-math} and DAPO~\citep{Yu2025DAPO} 
typically employ outcome reward modeling and operate without the utilization of expert trajectories.
We adopt \textbf{GRPO}, which eliminates the critic value function by employing 
group-normalized advantages $A_{i,t}$ derived from a group of $G$ self-generated trajectories. 
The objective maximizes the expected advantage subject to a KL-divergence constraint against the reference policy $\pi_{\text{ref}}$. 
To further enhance stability, we utilize \textbf{DAPO} as our primary RL algorithm. DAPO builds upon GRPO by introducing asymmetric clipping 
ratios $(\epsilon_{\text{low}},\epsilon_{\text{high}})$ and a dynamic difficulty sampling mechanism to filter extreme easy or hard prompts.

\noindent \textbf{Synchronized SFT-RL (Syn-SFT-RL)} 
paradigm attempts to merge the benefits of both SFT and RL by explicitly injecting expert trajectories into the RL optimization loop. 
We consider four representative baselines: (1) \textbf{LUFFY}~\citep{yan2025learningreasonoffpolicyguidance}, which jointly optimizes on-policy 
rollouts and off-policy expert trajectories using regularized importance shaping; (2) \textbf{SRFT}~\citep{Fu2025SRFT}, which balances SFT, 
off-policy RL, and likelihood-shaping losses via entropy-guided dynamic weights; and (3) \textbf{UPT}~\citep{lv2025unifiedviewlargelanguage}, 
which employs a hard gating mechanism to switch between SFT (stabilization) and GRPO (exploration) based on prompt-level reward thresholds.
(4) \textbf{Prefix-RFT}~\citep{Huang2025PrefixRFT}, which samples prefixes from expert trajectories to
guide model exploration, treating the composite sequence of an off-policy prefix and its on-policy continuation as a unified trajectory for
reinforcement updates.

\section{The Plasticity-Ceiling Framework}
\label{sec:framework} 
To systematically evaluate the trade-offs among post-training methods, 
we propose the {Plasticity-Ceiling} analytical framework. 
Unlike prior works that study SFT or RL scaling in isolation~\citep{ScaleLaw_downstream, rlscale},
our framework treats the SFT-then-RL pipeline as a unified continuum. This
allows us to quantify the respective contributions of the SFT and RL phases to
the overall post-training performance ceiling ($A_{\text{post}}$) in Def.~\ref{def:ceiling}.
\subsection{Decompose the Post-training Performance}
\label{sec:decomposition} 
Formally, we decompose the post-training performance
$P_{\text{post}}$ of the typical {SFT-then-RL} pipeline into three distinct
components based on the training stages:
\begin{equation}
    \label{eq:Apost_decompose}
    \small
    \begin{gathered}
        P_{\text{post}}(x_{\text{sft}},x_{\text{rl}})=P_{0}+\underbrace{(P_{\text{sft}}(x_{\text{sft}})-P_{0})}_{\text{SFT gain} ,\  \Delta P_{\text{sft}}(x_{\text{sft}})}\\
        \qquad \qquad \qquad \qquad +\underbrace{\
 (P_{\text{rl}}(x_{\text{rl}})-P_{\text{sft}}\left(
        x_{\text{sft}} \right) )}_{\text{RL gain} ,\  \Delta P_{\text{rl}}(x_{\text{rl}})}\\
        P_{\text{sft}}\left( x_{\text{sft}}=0 \right) =P_{0},\  P_{\text{rl}}(x_{\text{rl}}=0)=P_{\text{sft}}\left(
        x_{\text{sft}} \right) \,,
    \end{gathered}
\end{equation}
where $P_{0}$ denotes the base model's initial performance, and $x_{\text{sft}}$,
$x_{\text{rl}}$ denote the compute cost in FLOPs (see Appx.~\ref{appx:flops_est})
during the SFT and RL phases, respectively. This decomposition explicitly isolates
the performance contributors: $\Delta P_{\text{sft}}$ represents the gain
realized from SFT given cost $x_{\text{sft}}$, while $\Delta P_{\text{rl}}$ represents
the gain from RL given cost $x_{\text{rl}}$. Notably, Eq.~\ref{eq:Apost_decompose} reduces to Pure-SFT when $x_{\text{rl}}= 0$,
and to Pure-RL (including {Syn-SFT-RL} variants) when $x_{\text{sft}}= 0$.
 
\subsection{Ceiling and Plasticity}
\label{sec:dynamics}
Since RL is the final stage of the post-training pipeline, its scaling behavior determines the ultimate 
performance ceiling. To quantify this, we model the post-training performance $P_{\text{post}}$ as a 
function of RL compute $x_{\text{rl}}$ using a sigmoidal law~\citep{rlscale}. This law is initialized 
from $P_{\text{sft}}(x_{\text{sft}})$, which is the realized SFT outcome given the SFT compute $x_{\text{sft}}$:
\begin{equation}
    \small
    \label{eq:rl_scaling}
    \begin{aligned}
    P_{\text{post}}(x_{\text{sft}}, x_{\text{rl}}) = \ & (\epsilon +P_{\text{sft}}(x_{\text{sft}})) \\
    & + \frac{A_{\text{post}}- (\epsilon + P_{\text{sft}}(x_{\text{sft}}))}{1 + (x_{\text{rl}}/C_{\text{mid}_{\text{rl}}})^{-B_{\text{rl}}}}
    \end{aligned}
    \end{equation}
In the equation, the alignment offset $\epsilon \ge 0$ captures potential discontinuities during the phase transition, 
providing the flexibility to accommodate empirical misalignments between the SFT endpoint and the global RL scaling trend (see non-zero $\epsilon$ fitting cases in Figure~\ref{fig:s1k}).
The midpoint parameter $C_{\text{mid}_{\text{rl}}}$ measures the RL compute required to 
realize half of the gain term, $A_{\text{post}} - (\epsilon + P_{\text{sft}}(x_{\text{sft}}))$, while the scaling steepness $B_{\text{rl}}$ controls the rate of convergence.
On top of these parameters, this framework allows us to formalize two key metrics as follows:
\begin{definition}[\textbf{Asymptotic Ceiling}]
    \label{def:ceiling}
    For a given SFT compute allocation $x_{\text{sft}}$, the post-training ceiling $A_{\text{post}}$ defines the theoretical upper bound of performance achievable 
    in the limit of infinite RL computation. Formally, it is expressed as:
    \begin{equation}
        \label{eq:asymptotic_ceiling}
        \lim_{x_{\text{rl}} \to \infty} P_{\text{post}}(x_{\text{sft}}, x_{\text{rl}}) = A_{\text{post}}(x_{\text{sft}})\,.
    \end{equation}
\end{definition}

\begin{definition}[\textbf{RL Plasticity}]
    \label{def:plasticity}
    The RL plasticity quantifies the maximum effective headroom available for performance 
    improvement relative to the realized SFT baseline. Formally, it is 
    defined as the difference between the asymptotic post-training ceiling and the corresponding 
    SFT performance:
    \begin{equation}
        \label{eq:rl_plasticity}
        PL_{\text{rl}}(x_{\text{sft}}) = A_{\text{post}}(x_{\text{sft}}) - P_{\text{sft}}(x_{\text{sft}})\,.
    \end{equation}
\end{definition}
These definitions imply that both $A_{\text{post}}$ and $PL_{\text{rl}}$ are $x_{\text{sft}}$-dependent quantities, whose values vary with the allocated SFT compute.
In practice, $A_{\text{post}}$ is estimated using a robust estimator detailed in Appx.~\ref{appx:robust_curve_fit}, with comprehensive 
fitting results provided in Table~\ref{tab:fit_res_tab}. 

\noindent \textbf{Theoretical Implication.} 
Such framework reveals a fundamental insight:
optimizing the SFT solely improves the $P_{\text{sft}}$, but it may shrink the $PL_{\text{rl}}$
and thereby constrain $A_{\text{post}}$ if the SFT data is suboptimal (e.g., limited in scale). Thus, it is crucial to 
understand how $P_{\text{sft}}$, $PL_{\text{rl}}$ influence $A_{\text{post}}$ in the post-training.

\vspace{-0.5em}
\section{Experimental Setup}\label{sec:exp_setup}
\vspace{-0.5em}
To determine the optimal mechanism for utilizing expert trajectories, we organize experiments progressively to 
address three core research questions:

\noindent \textbf{RQ1: Paradigm Selection.} 
Among {Pure-RL}, {Pure-SFT}, {Syn-SFT-RL}, and {Sequential SFT-then-RL}, which paradigm establishes the most 
effective post-training baseline, and what are their characterizations? 

\noindent \textbf{RQ2: Optimal SFT-to-RL Transition.} Building upon the optimal paradigm identified in \textbf{RQ1}, 
what is the optimal time to transit to RL from SFT for a maximum final ceiling?

\noindent \textbf{RQ3: Data Properties (Scale \& Difficulty).} 
With the paradigm (\textbf{RQ1}) and optimal timing strategy (\textbf{RQ2}) established, what roles do data scale and 
difficulty play in maximizing the performance ceiling, and do they support or refute the ``Less is More'' hypothesis?

\vspace{-0.5em}
\subsection{Models and Data}
\noindent \textbf{Models.}
We primarily use Qwen2.5-7B~\citep{qwen2025qwen25technicalreport} in \S~\ref{sec:paradigm_compare} and \ref{sec:sft_then_rl_paradigm}, 
and Llama3.2-3B~\citep{llama32} in \S~\ref{sec:validate_llama} for cross-validation. In \S~\ref{sec:paradigm_compare}, 
we apply {Syn-SFT-RL} algorithms to Qwen2.5-Math-7B~\citep{qwen2_5_math} to further examine the influence of model priors.
\begin{figure*}[t] 
    \centering
    \includegraphics[width=\linewidth]{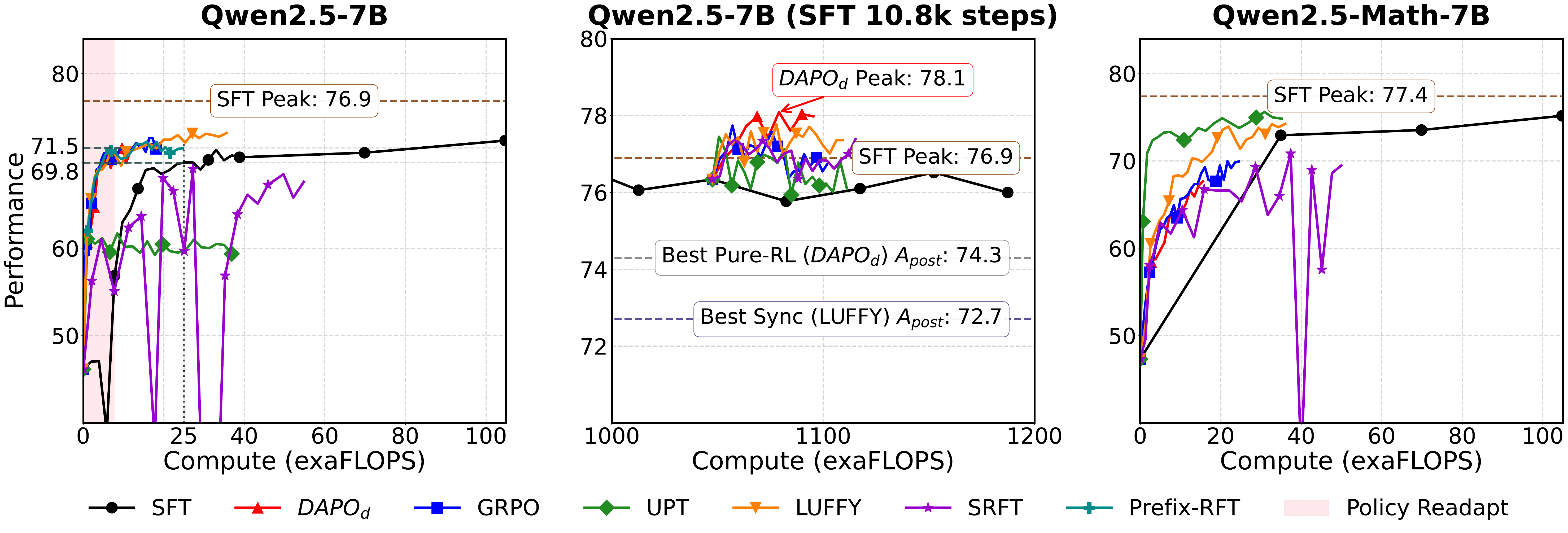}
    \caption{\textbf{The training dynamic comparision of post-training paradigms under different initializations.} 
(\textbf{Left}) Initializing from Qwen2.5-7B. {Early RL-like runs converge quickly (except unstable instances), while early SFT shows a mild performance disruption due to policy readaptation~\citep{chord-d}.} 
(\textbf{Middle}) Initializing from a saturated SFT checkpoint (10,800 steps on Qwen2.5-7B), where SFT-then-DAPO$_{d}$ outperforms other paradigms.
(\textbf{Right}) Initializing from Qwen2.5-Math-7B, where UPT and LUFFY demonstrate notable efficiency advantages.}
    \vspace{-0.5cm}
    \label{fig:paradigm_compare}
\end{figure*}

\noindent \textbf{Training Data.}
We construct SFT datasets of varying scales and difficulties by curating mathematical trajectories from distilled DeepSeek-R1~\citep{deepseek-r1} 
outputs~\citep{amthink1_4m, amthink40m}. The resulting datasets include the large-scale {SFT889K} with around 889K unique samples, 
three medium-scale variants controlled for difficulty ({Uniform/Easy/Hard102K}, refer to Table~\ref{tab:data_character} for the 
difficulty classification), and a held-out validation set {Val-199} with 199 prompt and trajectory pairs. To test data efficiency, 
we also include {S1K-1.1}~\citep{s1} (S1K for short), containing 1K high-quality R1-style trajectories.

For RL in the SFT-then-RL pipeline, we use {RL62K}, a filtered prompt set from Skywork-OR1-RL~\citep{skywork-or1}. 
For {Syn-SFT-RL} methods, we augment RL62K with expert trajectories in SFT889K to create {MIX37K}. 
Refer to Appx.~\ref{appx:sft_data_curate} for details.

\noindent \textbf{Benchmarks.}
To prevent data leakage, we filter out the benchmark prompts with over $0.8$ cosine 
similarity against our training set using Qwen3-8B-Embedding~\citep{qwen3embedding}.
We evaluate on the resulting 2,157 unique problems from the following cleaned 
benchmarks (counts denote original to cleaned):
{GSM8K}~\citep{cobbe2021gsm8k} (1319 to 1317),
{OlympiadBench}~\citep{olympiadbench} (675 to 291),
{Minerva}~\citep{minerva} (272 to 262),
{MATH}~\citep{lightman2023letsverifystepstep} (500 to 237),
and {AIME24/25}~\citep{numina_math_datasets} (30 to 25).
We report the average performance on these unique problems unless otherwise specified. Furthermore, we 
collect GPQA-Diamond (GPQA-D)~\citep{gpqa} (198 prompts) for graduate-level science and 
ARC-C~\citep{arc-c} (1,172 prompts) for robust open-domain reasoning to evaluate out-of-domain (OOD) 
generalization.

\subsection{Training and Evaluation}
\noindent \textbf{Training.} Our experiments include two primary paradigms:
(1) {Syn-SFT-RL}: we implement UPT~\citep{lv2025unifiedviewlargelanguage}, SRFT~\citep{Fu2025SRFT} and 
LUFFY~\citep{yan2025learningreasonoffpolicyguidance} using the UPT codebase~\cite{lv2025unifiedviewlargelanguage}.
Prefix-RFT~\citep{Huang2025PrefixRFT} is implemented using its official codebase.
(2) {Sequential SFT-then-RL}: we first fine-tune the base model on SFT data, then 
apply RL on the fine-tuned checkpoints. For comparison in §~\ref{sec:paradigm_compare}, 
we adopt GRPO and DAPO$_{d}$ (GRPO with dynamic difficulty sampling~\citep{Yu2025DAPO}) 
as the {Pure-RL} baseline. When implementing the {SFT-then-RL} pipeline in~\ref{sec:sft_then_rl_paradigm}, 
we use the enhanced DAPO$_{dc}$ in the RL phase, which further incorporates asymmetric ratio clipping 
into DAPO$_{d}$. See Appx.~\ref{appx:train_config} for implementation details.

\noindent \textbf{Evaluation.}
We report {pass@1} accuracy sampled with a temperature of 0.7 and top-p 1.0 to ensure generation diversity. For the smaller AIME24/25 datasets, we use {Avg@16} for robust estimation. All responses are generated with a maximum length of 8,192 tokens.

\section{Experimental Results}
\label{sec:experimental_results}
\subsection{Paradigms Comparison}
\label{sec:paradigm_compare}
To determine the optimal paradigm (\textbf{RQ1}), we systematically 
benchmark three approaches: {Pure-SFT}, {Pure-RL} (GRPO, DAPO$_{d}$), 
{Syn-SFT-RL} (LUFFY, SRFT, UPT, Prefix-RFT), and the {SFT-then-RL} pipeline. 
Following the data filtering setting of~\citet{yan2025learningreasonoffpolicyguidance},
all RL (or Syn-SFT-RL) runs utilize MIX37K, a subset of SFT889K. Our
preliminary experiments (Appx.~\ref{appx:paradigm_compare_sft889k}) confirm that 
training with these methods on the SFT889K leads to the suboptimal convergence.
Thus, MIX37K serves as a robust proxy to capture their performance upper bounds.
As shown in Figure~\ref{fig:paradigm_compare},  most of RL or Syn-SFT-RL 
methods typically saturate or destabilize within a single epoch, demonstrating that MIX37K 
suffices to capture their performance limits.

\noindent \textbf{Limitations of Syn-SFT-RL.} Contrasting prior claims~\citep{yan2025learningreasonoffpolicyguidance, lv2025unifiedviewlargelanguage, Fu2025SRFT},
our experiments reveal severe practical limitations in {Syn-SFT-RL}
methods, which exhibit training instability. For instance, SRFT shows performance
fluctuations with a standard deviation {2.6$\times$ higher} than the stable DAPO$_{d}$
baseline in Figure~\ref{fig:paradigm_compare} (Left) and fails to converge stably from a saturated SFT checkpoint (Figure~\ref{fig:paradigm_compare} Middle). 
Furthermore, they are highly sensitive to model priors. 
For instance, UPT achieves superior convergence on Qwen2.5-Math-7B, 
outperforming other RL variants (Figure~\ref{fig:paradigm_compare} Right).
However, this advantage vanishes on general-purpose Qwen2.5-7B, where 
it even underperforms vanilla GRPO (Figure~\ref{fig:paradigm_compare} Left).

\noindent \textbf{RL Variants Trade Ceiling for Efficiency.}
Pure-RL and stable Syn-SFT-RL methods exhibit a common trade-off: high initial efficiency but a limited ceiling.
As shown in the left panel of Figure~\ref{fig:paradigm_compare},
GRPO, DAPO$_d$, LUFFY, and Prefix-RFT rapidly reach around 71.5 points, outperforming Pure-SFT (69.8) within 25 exaFLOPs.
However, their performance quickly plateaus thereafter.
The middle panel of Figure~\ref{fig:paradigm_compare} shows that even with infinity RL compute,
DAPO$_d$ improves only by an additional 2.8 points to reach a ceiling of 74.3,
while LUFFY gains only 1.5 points to reach 72.7. This suggests that without a dedicated SFT phase,
the headroom for improvement remains structurally constrained in post-training. \looseness=-1

\noindent \textbf{SFT Foundation and Sequential RL Maximization.} 
In contrast, {Pure-SFT} demonstrates ``Slow but High'' scaling in 
the middle panel of Figure~\ref{fig:paradigm_compare}.
SFT achieves the continuous improvement through the extensive training and reach a peak of 76.9 points. 
It significantly surpasses the highest {Pure-RL} and {Syn-SFT-RL} methods' ceilings, 
which are 74.3 of DAPO$_d$ and 72.7 of LUFFY, respectively. 
Crucially, transitioning to RL after SFT saturation successfully unlocks further gains.
Specifically, SFT-then-DAPO$_\text{d}$ achieves the 
best performance with 78.1 points, extending the post-training performance frontier by  
synergizing the SFT performance with further RL improvement.

\noindent \textbf{Answer to RQ1: Sequential SFT-then-RL is the superior paradigm.}
Large-scale SFT provides the requisite robust foundation, and subsequent RL leverages this foundation to 
achieve the final performance frontier.

\begin{figure*}[t]
    \centering
    \includegraphics[width=1\linewidth]{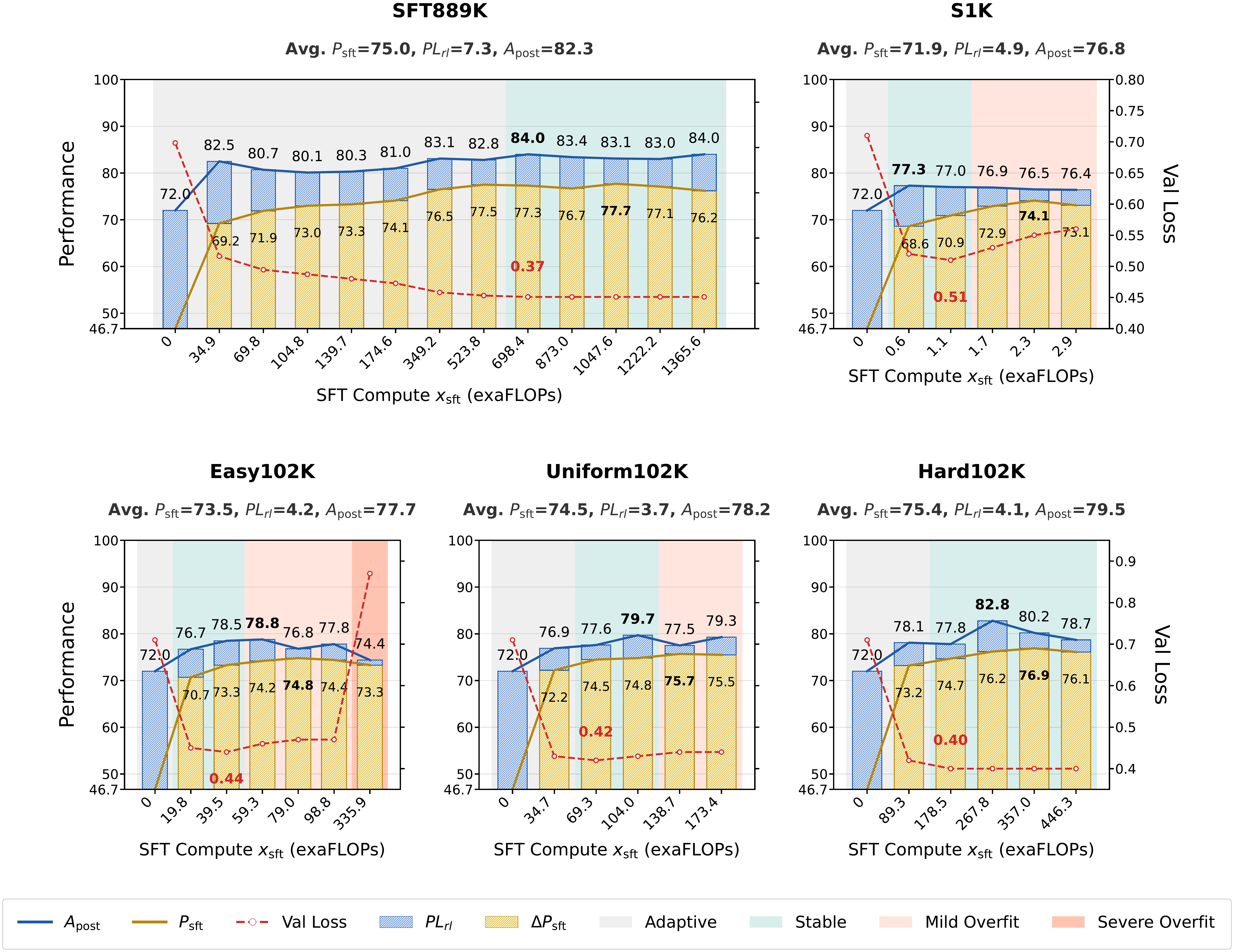}
    \caption{\textbf{SFT-then-RL scaling dynamics across data properties.}
The post-training ceiling ($A_{\text{post}}$) decomposes into SFT performance ($P_{\text{sft}}$) and RL plasticity ($PL_{\text{rl}}$), plotted against SFT compute $x_{\text{sft}}$ (exaFLOPs).
Background colors mark the four SFT regimes (\S~\ref{sec:compute_impact}), segmented by the SFT validation loss curves.
Note: Pure-RL ($x_{\text{sft}}=0$) and the Easy102K severe-overfitting checkpoint ($x_{\text{sft}}=335.9$) are excluded from the averaging statistics.
}
    \vspace{-0.5cm}
    \label{fig:vary_datasets}
\end{figure*}

\begin{figure*}[t] 
    \centering
    \includegraphics[width=0.9\linewidth]{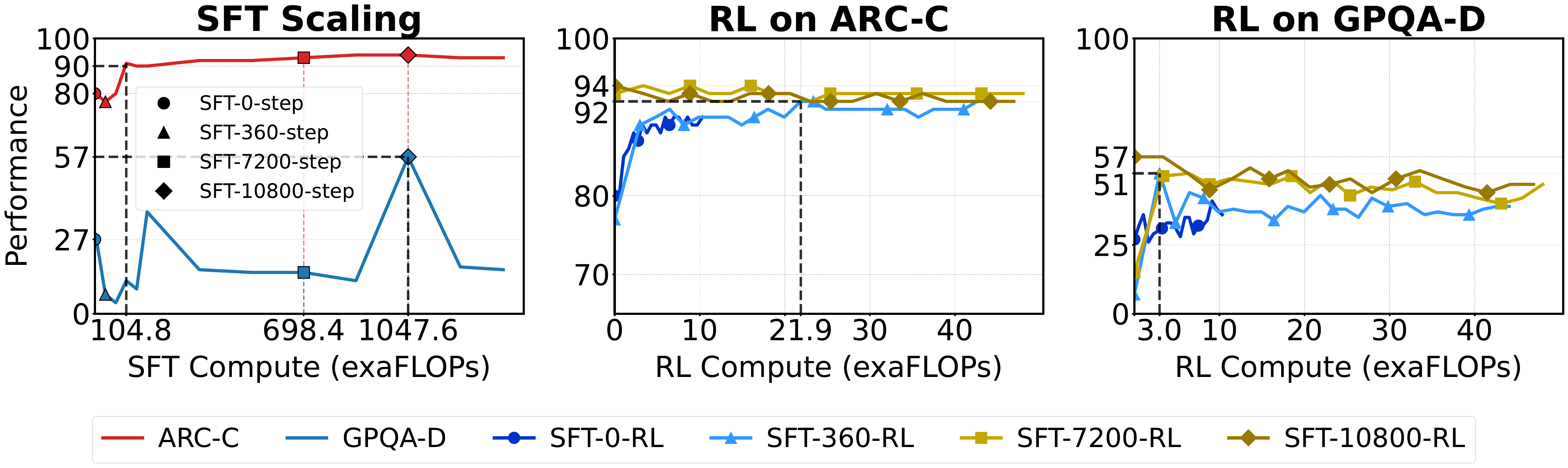}
    \vspace{-0.5em}
    \caption{\textbf{The OOD generalization analysis for SFT-then-RL on SFT889K.} 
    (\textbf{Left}) The performance dynamics on ARC-C and GPQA-D during SFT. 
    (\textbf{Middle\&Right}) The performance dynamics on ARC-C and GPQA-D during RL initiating from different SFT checkpoints.
    }
    \vspace{-0.5cm}
    \label{fig:ood_eval_sft_time}
\end{figure*}
\vspace{-0.2cm}

\subsection{SFT-then-RL Pipeline}
\label{sec:sft_then_rl_paradigm} 
Given SFT-then-RL's superiority (\S~\ref{sec:paradigm_compare}), 
we now examine the two determinants of its final ceiling ($A_{\text{post}}$): 
SFT compute allocation (\textbf{RQ2}) and data properties (\textbf{RQ3}). 
We first derive the optimal SFT-to-RL transition timing, then analyze the impact of data 
scale and difficulty.

\subsubsection{The Impact of SFT Compute Allocation}
\label{sec:compute_impact} 
Balancing realized SFT performance ($P_{\text{sft}}$) against preserving RL 
plasticity ($PL_{\text{rl}}$) is crucial for determining the optimal SFT-to-RL transition.

\noindent \textbf{SFT Regimes.}
To rigorously identify the optimal transition point, we temporally partition the SFT process based on the SFT validation 
loss $L(x_{\text{sft}})$. The entire trajectory is segmented based on the following mathematically defined regions:
\begin{equation}
    \small
    \begin{aligned}
    \mathcal{T}_{\text{stable}} &= \{\,x_{\text{sft}} \mid L(x_{\text{sft}}) \le (1+\delta)L_{\min}\,\}, \\
    \mathcal{T}_{\text{mild}}   &= \{x_{\text{sft}} \mid (1+\delta)L_{\min} < L(x_{\text{sft}}) \\
                                &\qquad < (1+\delta_2)\cdot L_{\min}\}, \\
    \mathcal{T}_{\text{severe}} &= \{\,x_{\text{sft}} \mid L(x_{\text{sft}}) \ge (1+\delta_2) \cdot L_{\min}\,\}.
    \end{aligned}
\end{equation}
where $L_{\min}$ is the global minimum validation loss observed during training, $(\delta, \delta_2)$ are tolerance 
thresholds being set as (0.02, 0.1) empirically. Therefore, we have four separable regimes characterizing the 
SFT convergence: 
\begin{itemize}[leftmargin=12pt, itemsep=0.45em, topsep=0.35em, parsep=0.15em]
    \item  \textbf{Adaptive Regime} ($\mathcal{P}_{\text{adapt}}$), where SFT is underfitting in the region.
            \begin{equation}\small
                \mathcal{P}_{\text{adapt}} = \{x_{\text{sft}} \mid x_{\text{sft}} < \min \mathcal{T}_{\text{stable}}\}
            \end{equation}
    \item  \textbf{Stable Regime} ($\mathcal{P}_{\text{stable}}$), where the validation loss saturates within a small tolerance threshold of 2\% (i.e., $\delta=0.02)$.
            \begin{equation}\small
                \mathcal{P}_{\text{stable}} = \mathcal{T}_{\text{stable}}
            \end{equation}
    \item \textbf{Mild Overfitting Regime} ($\mathcal{P}_{\text{mild}}$), where the region where loss rises slightly but remains below the 10\% tolerance, representing the ``risky sweet spot.''
            \begin{equation}\small
                \mathcal{P}_{\text{mild}} = \{x_{\text{sft}} \mid x_{\text{sft}} > \max \mathcal{T}_{\text{stable}} \text{ and } x_{\text{sft}} \in \mathcal{T}_{\text{mild}}\}
            \end{equation}
    \item \textbf{Severe Overfitting Regime} ($\mathcal{P}_{\text{severe}}$), where loss significantly diverges ($\ge 10\%$ rise when $\delta_2=0.1$), leading to rapid plasticity collapse (see Easy102K in Figure~\ref{fig:vary_datasets}).
            \begin{equation}\small
                \mathcal{P}_{\text{severe}} = \{x_{\text{sft}} \mid x_{\text{sft}} > \max \mathcal{T}_{\text{stable}} \text{ and } x_{\text{sft}} \in \mathcal{T}_{\text{severe}}\}
            \end{equation}
\end{itemize}

\noindent \textbf{The Dynamics of Post-training Ceiling.} The blue solid line in Figure~\ref{fig:vary_datasets}
illustrates how $A_{\text{post}}$ evolves across these regimes.
We observe that initiating RL prematurely during the {Adaptive Regime} is
consistently suboptimal because the model lacks foundational competence that
subsequent RL cannot fully recover. For instance, on SFT889K, switching early at
69.8 exaFLOPs yields a ceiling of only 80.7 points, whereas extending training
to the {Stable Regime} (698.4 exaFLOPs) boosts the ceiling to its peak of 84.0
points. Ideally, for most of SFT datasets (e.g., SFT889K, S1K, Uniform102K, Hard102K), the
{Stable Regime} aligns perfectly with peak performance. However, on Easy102K, the peak ceiling shifts 
into the {Mild Overfitting Regime}, indicating that a slightly delayed transition is acceptable and can 
even be beneficial due to $P_{\text{sft}}$ improvement.

Conversely, aggressively continuing SFT into the {Severe Overfitting Regime} is
detrimental. As demonstrated on Easy102K in Figure~\ref{fig:vary_datasets}, training
SFT to 335.9 exaFLOPs leads to a rapid decline in the final ceiling due to a collapse
in RL plasticity.

\noindent \textbf{Evaluation on OOD Generalization.}
Figure~\ref{fig:ood_eval_sft_time} illustrates the OOD performance of SFT889K checkpoints 
and their subsequent RL progress. The left panel demonstrates substantial SFT-driven gains: 
ARC-C improves by 10 points to surpass 90.0 with 104.8 exaFLOPs (1,080 steps), while GPQA-D increases by 
30 points to peak at 57.0 after 1047.6 exaFLOPs (10,800 steps). Contrary to claims that intensive SFT degrades 
generalization~\citep{scalpel_hammer,rl_razor}, our results prove it establishes a robust OOD 
reasoning foundation. We attribute this discrepancy to initialization, as prior studies 
evaluated instruction-tuned models rather than pre-trained base models.

The middle and right panels depict end-to-end performance as RL compute evolves. 
Transitioning to RL prematurely (SFT-360-RL, Adaptive Regime) yields significant RL-phase gains but 
suffers from the early convergence, restricting optimal scores to 92.0 on ARC-C and 51.0 on GPQA-D. 
Conversely, transitioning from a saturated SFT foundation (SFT-10800-RL, Stable Regime) 
provides marginal RL improvements but unlocks higher ultimate ceilings of 94.0 and 57.0, 
respectively. Thus, training SFT to saturation remains vital for maximizing OOD generalization.

\noindent \textbf{Answer to RQ2: Train SFT to Saturation.} 
Across both In-Domain (ID) and OOD tasks, the optimal strategy targets the \textbf{Stable Regime}, avoiding Severe Overfitting to preserve RL plasticity. 
While Mild Overfitting may be tolerable for select datasets, the Stable Regime provides a robust standard for maximizing the performance 
ceiling with scalable data.

\vspace{-0.5em}
\subsubsection{The Impact of SFT Data Properties}
\label{sec:data_impact} Data scale and difficulty are critical determinants of
the quality of the SFT prior. We now focus on investigating how
these data properties influence the asymptotic post-training
performance ceiling. While we focus on representative scales here, extended results 
across additional data granularities are provided in Appx.~\ref{appx:further_data_scale_study}.

\noindent \textbf{Larger Scale Begets Higher Ceiling.} Comparing datasets of varying scales
(S1K, Uniform102K, and SFT889K in Figure~\ref{fig:vary_datasets}) reveals that
while minimal data can achieve rapid initial SFT gains, extensive data scale is indispensable
for reaching a higher post-training ceiling. Initially, small-scale data exhibits
deceptive efficiency: S1K achieves an SFT performance of 74.1
points using only 2.3 exaFLOPs, matching the similar performance level that requires 69.3
exaFLOPs on Uniform102K ($P_{\text{sft}}$=74.5) and 174.6 exaFLOPs on SFT889K ($P_{\text{sft}}$=74.1). 

However, this efficiency proves
to be unsustainable. The realized SFT performance $P_{\text{sft}}$ of S1K
saturates prematurely at this level. In contrast, Uniform102K and SFT889K
continue to improve with additional compute. 
Uniform102K reaches SFT performances of 74.8 at 104.0 exaFLOPs
while SFT889K reaches 77.3 at 698.4 exaFLOPs, which thereby establish the 
superior foundation for the subsequent
RL phase. Crucially, the large-scale SFT also preserves greater RL plasticity.
SFT889K maintains an average $PL_{\text{rl}}$ of 7.3, exceeding both S1K and Uniform102K
by 2.4 and 3.7 points. Consequently, by leveraging both higher realized SFT performance 
$P_{\text{sft}}$ and enhanced RL plasticity $PL_{\text{rl}}$, the large-scale SFT unlocks a
significantly higher post-training ceiling. \looseness -1

\noindent \textbf{Harder Data Elevates the Ceiling.} Controlling the dataset size as 102K, 
we investigate how trajectory difficulty affects performance by comparing Easy102K, Uniform102K, 
and Hard102K. As shown in the bottom row of Figure~\ref{fig:vary_datasets}, training on harder 
trajectories leads to higher returns. Specifically, Hard102K achieves a higher SFT 
performance ($P_{\text{sft}}$) of 75.4 points in average, outperforming Easy102K (73.5) and Uniform102K 
(74.6) by 1.9 and 0.8 percentage points, respectively. At the same time, Hard102K maintains a 
similar RL plasticity ($PL_{\text{rl}}$) to its 102K counterparts, with a score of 4.1 points. 
This combination, which includes higher SFT performance and decent RL plasticity, makes harder 
data the superior choice for maximizing the post-training ceiling.

\noindent \textbf{Minimum Validation Loss as an Indicator.} 
Across diverse SFT data configurations, the minimum validation loss shows a strong negative 
correlation with $A_{\text{post}}$ (Pearson $r$=-0.92, Figure~\ref{fig:correlation}). Hence, 
it serves as an efficient a priori indicator: a lower loss reliably predicts a higher post-training 
ceiling in the SFT-then-RL pipeline, bypassing expensive RL runs.

\noindent \textbf{Answer to RQ3: Scale Dominates, Difficulty Optimizes.} 
Refuting ``Less is More'', we establish Data \textbf{Scale} as the primary factor to improve the post-training ceiling, while \textbf{Difficulty} 
acts as a multiplier. Harder trajectories are helpful when the data scale is limited. 
Thus, scaling must prioritize volume before difficulty, with the final ceiling reliably indicated by the minimum SFT validation loss. \looseness -1
\begin{figure}[t]
    \centering
    \includegraphics[width=1\linewidth]{
        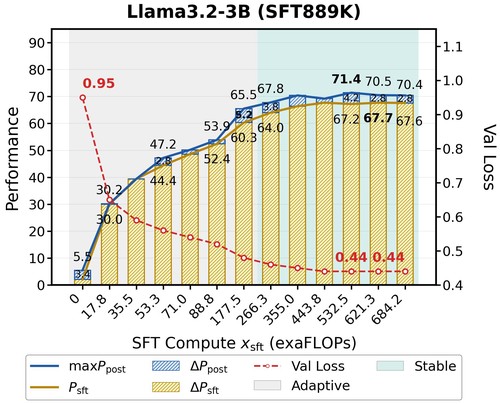
    }
    \caption{
\textbf{The analysis of the max post-training performance $\max P_{\text{post}}$ when performing SFT-then-RL on Llama3.2-3B.}
 SFT is performed on SFT889K. {The Stable Regime begets a higher $\max P_{\text{post}}$ result.}}
    \label{fig:llama_sft_then_rl}
    \vspace{-0.5cm}
\end{figure}

\vspace{-0.5em}
\subsection{The Validation on Llama3.2-3B}
~\label{sec:validate_llama} 
We validate our findings on Llama3.2-3B ~\citep{llama32}
to for cross-model generalization. To prioritize
practical relevance, we report the maximum achieved post-training performance ($\max P_{\text{post}}$) instead of the theoretical ceiling ($A_{\text{post}}$), within 200-step
RL training. 

Llama3.2-3B results strongly corroborate our Qwen2.5-7B findings:
(1) \textbf{Paradigm:} As shown in Table~\ref{tab:llama_validate}, 
SFT-then-RL pipeline remains strictly superior, achieving 
68.7 points. It far surpasses Pure-RL or Syn-SFT-RL baselines. 
(2) \textbf{Compute Allocation:} Consistent with Figure~\ref{fig:vary_datasets}, 
Figure~\ref{fig:llama_sft_then_rl} confirms that for smaller models, 
transitioning to RL from the {Stable Regime} is critical. Specifically, the Stable Regime 
achieves a peak $\max P_{\text{post}}$ of 70.4 at $x_{\text{sft}}$=532.5 exaFLOPs, 
outperforming the Adaptive Regime's best (65.5 points at $x_{\text{sft}}$=177.5) by 5.9 points.
(3) \textbf{Data Properties:}
Table~\ref{tab:llama_validate} confirms that data scale remains the primary determinant of the ceiling 
($\max P_{\text{post}}$ ranges from 25.0 on S1K to 70.4 on SFT889K), 
while difficulty acts as a secondary multiplier (Hard102K leads at 56.4 among 102K variants). Crucially, the minimum SFT validation loss retains 
its strong relevance with the final reasoning ceiling (Pearson $r$=-0.96).
\vspace{-0.5em}
\begin{table}[!htbp]
\caption{\textbf{Llama3.2-3B validation results.} 
We report the maximum post-training performance ($\max P_{\text{post}}$) 
and minimum SFT validation loss (Min. Val Loss). Best results are {bolded}.
For all SFT-then-RL experiments, we transition to RL at the checkpoint with 
Min. Val Loss (details in Appx.~\ref{appx:sft_config}).}  
\vspace{-0.5em}
\label{tab:llama_validate}
\centering
\resizebox{1.0\columnwidth}{!}{
\begin{tabular}{@{}cccc@{}}
\toprule
	{Methods}         & {Paradigm}             & {$\max P_{\text{post}}$} & {Min. Val Loss} \\ \midrule
Llama3.2-3B              & -                           & 2.1           & -                         \\ \midrule
DAPO$_d$                   & Pure RL                      & 3.0           & -                         \\
UPT                      & \multirow{2}{*}{Syn-SFT-RL}  & 12.2          & -                         \\
LUFFY                    &                              & 8.5           & -                         \\
S1K                      & \multirow{5}{*}{SFT}        & 24.2          & 0.71                       \\
Easy102K                 &                              & 52.4          & 0.59                      \\
Uniform102K              &                              & 53.6          & 0.54                      \\
Hard102K                 &                              & 55.6          & 0.50                      \\
SFT889K                  &                              & 67.7          & \textbf{0.44}             \\
SFT889K$\rightarrow$  DAPO$_d$      & SFT-then-RL                  & \textbf{68.7}          & -                         \\ \midrule
S1K $\rightarrow$  DAPO$_{dc}$         & \multirow{5}{*}{SFT-then-RL} & 25.0          & -                         \\
Easy102K $\rightarrow$  DAPO$_{dc}$    &                              & 53.9          & -                         \\
Uniform102K $\rightarrow$  DAPO$_{dc}$ &                              & 55.3          & -                         \\
Hard102K $\rightarrow$ DAPO$_{dc}$      &                             & 56.4          & -                         \\
SFT889K $\rightarrow$  DAPO$_{dc}$       &                              & \textbf{70.4} & -                         \\ \bottomrule
\end{tabular}
} 
\vspace{-1.0em}
\end{table}

\section{Conclusions}~\label{sec:conclusion}
This work presents the {\textbf{Plasticity-Ceiling Framework}} for optimizing expert trajectory utilization, 
formalizing the trade-off between supervised fine-tuning performance (\(P_{\text{sft}}\)) and reinforcement learning 
plasticity (\(PL_{\text{rl}}\)). We derive three core principles for effective scaling: 
(1) The sequential \textbf{SFT-then-RL pipeline} outperforms alternative paradigms in approaching the post-training performance ceiling. 
(2) Within this pipeline, RL should be initiated at \textbf{SFT saturation}, a point reliably predicted by validation 
loss minimization. 
(3) \textbf{SFT data scale} primarily determines the performance ceiling, and trajectory difficulty 
further optimizes the ceiling when data is limited. Together, these findings transform expert trajectory optimization from 
empirical guesswork into a systematic and predictable process, establishing a rigorous standard for maximizing mathematical 
reasoning model performance.

\section*{Limitations}
Our findings are limited to mathematical reasoning with R1-style expert trajectories. 
Whether the Plasticity-Ceiling Framework generalizes to other verifiable tasks (e.g., logic, coding), 
non-verifiable tasks (e.g., creative writing), or alternative trajectory styles (e.g., Gemini-style, human-annotated) 
remains an open question for future work.

\vspace{-0.5em}
\section*{Ethical Considerations}
\vspace{-0.5em}
All artifacts utilized in this work, 
including training datasets, evaluation 
benchmarks, and models, are open-sourced
and governed by their respective licenses, 
such as MIT, Apache-2.0, CC-BY-NC-4.0, 
CC-BY-SA-4.0, and the Llama 3.2 Community License.

\vspace{-0.5em}
\section*{Acknowledgments}
\vspace{-0.5em}
This work was supported in part by the National Natural Science Foundation of China (NSFC) under
No. 62576285, Research Center for Industries of the Future (RCIF) at Westlake University,
and Westlake Education Foundation.

\bibliography{custom}

\clearpage
\appendix
\section{Experimental Platforms}
\label{appx:training_platform}
All SFT experiments in ~\S~\ref{sec:paradigm_compare} run on 16 GPUs; RL and Syn-SFT-RL experiments in ~\S~\ref{sec:paradigm_compare} are implemented on 8 GPUs, and RL experiments in ~\S~\ref{sec:sft_then_rl_paradigm} are conducted on 128 Ascend 910B NPUs.

\section{Post-training Paradigms}
\label{appx:post_train_algos}
We summarize the details of algorithmic foundations used in our
study: (1) standard supervised fine-tuning (SFT), (2) reinforcement learning (GRPO
and DAPO), and (3) synchronized SFT-RL (Syn-SFT-RL) fusion methods used as single-stage
baselines.
\subsection{SFT} 
SFT tunes the policy $\pi_{\theta}$ via imitation learning using the answer and expert
trajectory pair $(\mathbf{q},\bm{\tau})$ in the SFT dataset $\mathcal{D}_{\text{SFT}}$:
    \begin{equation}
        \label{eq:sft_obj}
        \begin{split}
        \textstyle \mathcal{J}_{\text{SFT}}(\theta) \!=\! {}& -\mathbb{E}_{(\mathbf{q},\! \bm{\tau}) \sim \mathcal{D}_{\text{SFT}}}\!\Bigl[ \\
        & \textstyle \sum_{t=1}^{|\bm{\tau}|}\!\log \pi_{\theta}(\bm{\tau}_{t}\!\mid\!\mathbf{q},\!\bm{\tau}_{<t}) \Bigr]
        \end{split}
    \end{equation}
This paradigm reliably instills instruction-following and basic reasoning capabilities in the model~\citep{NEURIPS2022_b1efde53,deepseek-r1}.

\subsection{RL} 
RL extends the model beyond imitation by optimizing reward-guided
exploration. GRPO~\citep{deepseek-math} and DAPO~\citep{Yu2025DAPO} are two widely-used
RL algorithms.

\noindent \textbf{GRPO} replaces a critic with a group-normalized advantage (${A}_{i,t}$). For
each query-answer pair $(\textbf{q},\textbf{a})$ in dataset
$\mathcal{D_{\text{RL}}}$, GRPO samples $G$ response trajectories $\{\bm{\tau}
_{i}\}_{i=1}^{G}$ based on the old policy $\pi_{\theta_{\text{old}}}$. Each trajectory receives a rule-derived reward score ${R}_{i}$. The group-normalized advantage is computed as:
\begin{equation}
    \small
    \label{eq:group_level_adv}A_{i,t}= \frac{R_{i}- \text{mean}(\{R_{j}\}_{j=1}^{G})}{\text{std}(\{R_{j}\}_{j=1}^{G})}
\end{equation}
With the advantage, GRPO aims to maximize the expected advantage while regularizing the policy towards a reference policy $\pi_{\text{ref}}$ via the KL divergence term $\beta \cdot \mathbb{D}_{\text{KL}}\left[ \pi_{\theta}||\pi_{\text{ref}}\right]$. The policy loss $\mathcal{J}_{\text{GRPO}}{(\theta)}$ is:
\begin{equation}
    \small
    \label{eq:grpo_obj}
    \begin{aligned}
        \mathcal{J}_{\text{GRPO}}(\theta) &= \mathbb{E}_{(\mathbf{q}, \mathbf{a})\sim \mathcal{D}_{\text{RL}}, \{\bm{\tau}_i\}_{i=1}^{G}\sim\pi_{\theta_{\text{old}}}} \\
        & \left[ \frac{\sum_{i=1}^{G}\sum_{t=1}^{|\bm{\tau}_{i}|}\min \left( r_{i}^{t}A_{i,t}, C_{i}^{t}A_{i,t}\right)}{\sum_{i=1}^{G}|\bm{\tau}_{i}|} \right] \\
        & - \beta \cdot \mathbb{D}_{\text{KL}}\left[ \pi_{\theta}||\pi_{\text{ref}}\right]
    \end{aligned}
\end{equation}
where $r_{i}^{t}=\frac{\pi_{\theta}( \bm{\tau}_{i,t}\mid \textbf{q}_{},
\bm{\tau}_{i,<t})}{\pi_{\theta_{old}}( \bm{\tau}_{i,t}\mid \textbf{q}_{},
\bm{\tau}_{i,<t})}$ represents the importance ratio between the new and old policies
for token $\bm{\tau}_{i,t}$. Its clipped counterpart, $C_{i}^{t}=\text{clip}( r_{i}^{t}
,\ 1-\epsilon ,\ 1+\epsilon )$, confines the policy update within a trust
region, preventing excessively large and destabilizing policy updates.

\noindent \textbf{DAPO}~\citep{Yu2025DAPO} further stabilizes training via asymmetric clipping $(\epsilon
_{\text{low}},\epsilon_{\text{high}})$ and dynamically filter the prompts with all correct or wrong on-policy generations. We adopt DAPO as our primary RL algorithm due to its robustness on
mathematical reasoning tasks.

\subsection{Syn-SFT-RL}~\label{sec:Syn-SFT-RL-preliminary}
The Syn-SFT-RL paradigm merges SFT and RL by injecting expert trajectories into the optimization
loop. We introduce four typical algorithms: LUFFY~\citep{yan2025learningreasonoffpolicyguidance}, 
SRFT~\citep{Fu2025SRFT}, UPT~\citep{lv2025unifiedviewlargelanguage} and 
Prefix-RFT~\cite{Huang2025PrefixRFT}.

\paragraph{LUFFY} modifies the $\mathcal{J}_{\text{GRPO}}(\theta)$ in Eq.~\ref{eq:grpo_obj} by jointly optimizing on-policy trajectories and off-policy ones.
It removes both the KL regularization and importance-ratio clipping, and aggregates token-level advantages over a mixture of SFT and RL data. The mixture dataset $\mathcal{D}_{\text{MIX}}$ contains triplets $(\bm{q}, \{\bm{\tau}_{j}\}_{j=1}^{N},\bm{a})$ with the prompt $\bm{q}$, $N$ expert trajectories $\{\bm{\tau}_{j}\}_{j=1}^{N}$ ($N$=1 as the official setup), and  answer $\bm{a}$. Hence, LUFFY's loss is formalized as:
\begin{equation}
\label{eq:luffy_obj}
\small
\begin{aligned}
\mathcal{J}_{\text{LUFFY}}(\theta)
&=
\mathbb{E}_{\substack{
(\bm{q},\{\bm{\tau}_j\}_{j=1}^{N},\bm{a}) \sim \mathcal{D}_{\mathrm{MIX}}\\
\{\bm{\tau}_i\}_{i=1}^{G} \sim \pi_{\theta_{\mathrm{old}}}
}}
\Biggl[
\frac{1}{Z}
\sum_{j=1}^{N}
\sum_{t=1}^{|\bm{\tau}_j|}
\hat{r}_{j}^{t}\,\hat{A}_{j,t}
\\
&\qquad\qquad\qquad
+
\frac{1}{Z}
\sum_{i=1}^{G}
\sum_{t=1}^{|\bm{\tau}_i|}
r_{i}^{t}\,\hat{A}_{i,t}
\Biggr],
\end{aligned}
\end{equation}
where $Z = \sum_{j=1}^{N}|\bm{\tau}_j| + \sum_{i=1}^{G}|\bm{\tau}_i|$ normalizes over all tokens, and the mixed advantages are computed without normalization:
\begin{equation}
\label{eq:luffy_adv}
\hat{A}_{i,t}
=
R_{i}
-
\text{mean}\!\left(
\{R_{j}\}_{j=1}^{N}
\cup
\{R_{i}\}_{i=1}^{G}
\right),
\end{equation}
To avoid entropy collapse on off-policy data, LUFFY further applies regularized importance shaping, which transforms the importance ratio $r_{j}^{t}$ to $\hat{r}_{j}^{t} = r_{j}^{t} / (r_{j}^{t} + \gamma)$ with a small constant $\gamma = 0.1$.

\paragraph{SRFT} combines four components: (i) the standard SFT loss $\mathcal{J}_{\text{SFT}}$ in Eq.~\ref{eq:sft_obj},
(ii) the off-policy loss $\mathcal{J}_{\text{off}}$ from LUFFY (the first term in Eq.~\ref{eq:luffy_obj}),
and (iii) on-policy objectives for positive and negative trajectories in Eq.~\ref{eq:pos_neg_obj}.
For $M$ on-policy positive rollouts $\{\bm{\tau}_i^+\}_{i=1}^{M}$ and $G-M$ on-policy negative ones $\{\bm{\tau}_j^-\}_{j=1}^{G-M}$, SRFT maximizes the likelihood of positive trajectories while suppressing that of negative ones:
\begin{equation}
\label{eq:pos_neg_obj}
\begin{aligned}
\mathcal{J}_{\text{pos}}(\theta)
&=
-\mathbb{E}\Bigl[
\sum_{t=1}^{|\tau_i^+|}
\log \pi_{\theta}\bigl(\tau_{i,t}^{+}\mid \mathbf{q}_i,\tau_{i,<t}^{+}\bigr)
\Bigr],\\
\mathcal{J}_{\text{neg}}(\theta)
&=
\phantom{-}\mathbb{E}\Bigl[
\sum_{t=1}^{|\tau_j^-|}
\log \pi_{\theta}\bigl(\tau_{j,t}^{-}\mid \mathbf{q}_j,\tau_{j,<t}^{-}\bigr)
\Bigr].
\end{aligned}
\end{equation}
The final SRFT objective uses entropy-guided dynamic weights:
\begin{equation}
\label{eq:srft_obj}
\begin{aligned}
\mathcal{J}_{\text{SRFT}}
&=
w_{1}\,\mathcal{J}_{\text{SFT}}
+
\mathcal{J}_{\text{off}}
+
w_{2}\,\mathcal{J}_{\text{pos}}
+
\mathcal{J}_{\text{neg}},
\\
w_{1}
&=
0.5\cdot\text{stop\_grad}\bigl(
e^{-\mathcal{H}(\pi_{\theta})}
\bigr),
\\
w_{2}
&=
0.1\cdot\text{stop\_grad}\bigl(
e^{\mathcal{H}(\pi_{\theta})}
\bigr).
\end{aligned}
\end{equation}
where $\mathcal{H}(\pi_{\theta})$ denotes the policy entropy and \texttt{stop\_grad} prevents gradients from flowing through the weights.

\paragraph{UPT} employs a hard gating mechanism to mix SFT and RL.
Let $p$ denote the average reward over the trajectories sampled for the current prompt $\textbf{q}$, and $\gamma$ be a threshold.
UPT defines a mixed loss
\begin{equation}
\label{eq:upt_mixed}
\mathcal{J}_{\text{UPT}}
=
f_{p}\,\mathcal{J}_{\text{SFT}}
+
g_{p}\,\mathcal{J}_{\text{GRPO}},
\end{equation}
where $(f_p,g_p)$ are determined by $p$:
\begin{equation}
\label{eq:upt_gate}
(f_p,g_p)
=
\begin{cases}
(1,0), & p \le \gamma,\\
(0,1), & p > \gamma .
\end{cases}
\end{equation}
When the model performs poorly on a prompt ($p \le \gamma$), the gate prioritizes SFT-style imitation.
Once the reward exceeds the threshold ($p > \gamma$), the gate switches to pure GRPO optimization to focus on exploration.

\paragraph{Prefix-RFT} blends SFT and RL by utilizing sampled expert prefixes to 
guide online exploration. For a given prompt $\bm{q}$ and an expert trajectory $\bm{\tau}^*$, a 
prefix sequence of length $l$ is sampled. To mitigate position bias and introduce curriculum 
learning, the prefix length ratio is governed by a cosine decay scheduler whose lower bound 
gradually decays toward zero. The current policy $\pi_{\theta_{\text{old}}}$ then generates a 
continuation $\bm{\tau}^{\text{cont}}$ to complete the response, yielding a composite hybrid 
trajectory $\bm{\tau}_{\text{hybrid}}$. The model optimizes over a mixture of these hybrid sequences 
and standard on-policy rollouts using the GRPO objective:
\begin{equation}
\label{eq:prefix_rft_obj}
\small
\mathcal{J}_{\text{Prefix-RFT}}(\theta) = \mathcal{J}_{\text{GRPO}}\left(\theta \mid \{\bm{\tau}_{\text{hybrid}}\} \cup \{\bm{\tau}_{\text{on-policy}}\}\right)
\end{equation}
To prevent off-policy demonstration gradients from dominating the updates, Prefix-RFT 
incorporates an entropy-based clipping mechanism. This strategy zeroes out the advantages 
for all but the top-$k$\% high-entropy demonstration tokens, thereby concentrating optimization 
on critical uncertain junctures and ensuring stable training.

\section{Training Configuration}
\label{appx:train_config}
\subsection{SFT}
\label{appx:sft_config}
We train SFT889K and all 102K variants with batch size of 512 and learning rate 1e-5 for 8 and 9 epochs, respectively.
To study severe overfitting, we continue training Easy102K up to 6{,}120 steps (335.9 exaFLOPs).
Checkpoints are saved every 360 steps (0.2 epochs for SFT889K, 1.8 epochs for 102K variants).
For S1K, we follow the official setup: batch size 16, learning rate 1e-5, weight decay 1e-4, and train for 5 epochs, with checkpoints saved every 62 steps (1 epoch).

Specifically, for the SFT-then-RL experiments in Table~\ref{tab:llama_validate}, we initialize RL from 
the checkpoint with minimum SFT validation loss, as these checkpoints typically reside in the Stable Regime and are more likely to achieve the post-training ceiling for the given SFT data.
Concretely, we use step 124 (2 epochs) for S1K, step 1800 (9 epochs) for all three 102K variants (Easy102K, Uniform102K, Hard102K), and step 14080 (7.8 epochs) for SFT889K.

\subsection{RL Practice}
\label{appx:rl_practice}
All RL and Syn-SFT-RL runs optimize a binary correctness reward (1 if correct, 0 otherwise), 
This correctness is verified using a
script powered by Math-Verify~\citep{mathverify2024}.
Furthermore, token-level loss aggregation is
uniformly applied across all runs.

\subsubsection{RL in Paradigms Comparison}~\label{appx:rl_implement}
We summarize the RL configuration for the training in \S~\ref{sec:paradigm_compare}, including the Pure-RL (GRPO and $\text{DAPO}_{d}$), and Syn-SFT-RL (LUFFY, SRFT, UPT, Prefix-RFT). 
The shared training hyperparameters for Pure-RL and Syn-SFT-RL methods are summarized in Table~\ref{tab:rl_config}. Unless specified, all algorithms use this default configuration.

\paragraph{Pure-RL.}
GRPO and $\text{DAPO}_{d}$ serve as Pure-RL baselines. $\text{DAPO}_{d}$ adds the dynamic difficulty sampling strategy~\citep{Yu2025DAPO} on GRPO. For $\text{DAPO}_{d}$, the dynamic difficulty sampling strategy employs a batch size of 128 responses per inference round, and the asymmetric clipping ratio strategy is not applied.

\begin{table}[htbp]
\centering
\small
\caption{\textbf{Shared training hyperparameters} for GRPO, $\text{DAPO}_{d}$, LUFFY, SRFT, UPT and Prefix-RFT in \S~\ref{sec:paradigm_compare}.}
\begin{tabular}{cc}
\toprule
Hyperparameter & Default value \\
\midrule
Batch size & 128 \\
Update batch size & 64 \\
Rollout number & 8 \\
Max prompt length & 1,024 \\
Max response length & 8,192 \\
Learning rate & $1\mathrm{e}{-6}$ \\
Entropy coefficient & 0.001 \\
KL loss term & removed \\
Std.\ in group advantage norm & removed \\
\bottomrule
\end{tabular}
\label{tab:rl_config}
\end{table}

\paragraph{Syn-SFT-RL.}
We adopt training configurations from the Unify-Post-Training codebase~\citep{lv2025unifiedviewlargelanguage}, with UPT utilizing a specific learning rate of 5e-6. 
For rollout generation, UPT adaptively allocates up to 8 trajectories between on-policy and off-policy samples. 
Conversely, LUFFY~\citep{yan2025learningreasonoffpolicyguidance} and SRFT~\citep{Fu2025SRFT} maintain a fixed 7:1 ratio of on-policy to expert trajectories per prompt. 

Following the similar setting, Prefix-RFT~\citep{Huang2025PrefixRFT} generates 8 trajectories per prompt, comprising 7 standard on-policy rollouts (empty prefixes) and 
1 hybrid rollout guided by a sampled expert prefix. Specifically, the expert prefix length is dynamically sampled via a Beta distribution, where the lower bound follows a 
cosine decay schedule from 95\% to 5\% of the demonstration length throughout training, while the upper bound remains constant at 95\%. Additionally, Prefix-RFT incorporates 
an entropy-based clipping mechanism that zeroes out advantages for the 80\% lowest-entropy demonstration tokens, concentrating optimization on high-uncertainty regions.

The maximum trajectory length for all Syn-SFT-RL algorithms is set to 8,192, 
as consistently adopted in prior works~\citep{yan2025learningreasonoffpolicyguidance, Fu2025SRFT, lv2025unifiedviewlargelanguage,Huang2025PrefixRFT}. 
Other shared hyperparameters are configured as detailed in Table~\ref{tab:rl_config}.

\subsubsection{RL in SFT-then-RL pipeline}
Recognizing the superiority of $\text{DAPO}_{d}$ when starting from an SFT checkpoint 
in \S~\ref{sec:paradigm_compare}, we further improve $\text{DAPO}_{d}$ to $\text{DAPO}_{dc}$ as the RL method 
in the following {SFT-then-RL} experiments in \S~\ref{sec:sft_then_rl_paradigm} and \S~\ref{sec:validate_llama}. 
$\text{DAPO}_{dc}$ adds the asymmetric clipping ratio strategy to $\text{DAPO}_{d}$, 
setting $(\epsilon_{\text{high}}, \epsilon_{\text{low}})=(0.28, 0.2)$. In $\text{DAPO}_{dc}$, each 
inference round uses 128 responses for dynamic difficulty sampling. The batch size and update batch size are 64, 
the learning rate is 1e-6, and the rollout number is 8.  Maximum lengths for prompt and response are 1,024 and 8,192, respectively. Furthermore, the entropy and KL term coefficients are 0, and the group advantage normalization is enabled. 
\begin{table*}[t] 
\centering
\caption{\textbf{Statistical summary of the constructed SFT datasets.}
It details average prompt/response lengths, logical step 
counts (derived via semantic analysis in Appx.~\ref{appx:sft_data_curate}), and Win Rates (WR) across 
different DeepSeek model sizes. 
These metrics confirm the intended difficulty stratification, 
distinguishing the complexity levels of Easy, Uniform, and Hard 
subsets.
\textbf{Note:} Avg.~logical steps are computed on 10K randomly sampled instances, while other metrics are computed on the full dataset.}
\label{tab:data_character}
\resizebox{0.85\textwidth}{!}{%
\begin{tabular}{@{}ccccccc@{}}
\toprule
Dataset & Avg. prompt length & Avg. response length & Avg. logical steps & WR (1.5B) & WR (7B) & WR (671B) \\ \midrule
Easy102K    & 64  & 2253 & 10 & 1    & 0.98 & 0.99 \\
Uniform102K & 74  & 3673 & -  & 0.79 & 0.84 & 0.87 \\
Hard102K    & 101 & 8532 & 13 & 0.09 & 0.30 & 0.41 \\
SFT889K     & 78  & 3693 & -  & -    & -    & -    \\
S1K         & 127 & 9884 & -  & -    & -    & -    \\
Uniform220K & 74  & 3674 & -  & -    & -    & -    \\
Uniform440K & 74  & 3671 & -  & -    & -    & -    \\ \bottomrule
\end{tabular}%
}
\end{table*}

\section{Evaluation Configuration} 
\paragraph{ID Evaluation}
During evaluation, we find that Math-Verify may fail to parse some semantically correct answers, leading to missed positives.
To reduce this bias, we use two complementary evaluators: a \emph{strict} evaluator based on Math-Verify alone, and a \emph{lenient} evaluator that additionally allows a fuzzy string match when symbolic verification fails.
For instances on which the two evaluators disagree, we further use an LLM judge with the prompt in Table~\ref{tab:llm_as_judge} for adjudication.
The final reported results are based on this two-stage verification procedure.

\paragraph{OOD Evaluation}
Manual inspection of model outputs on GPQA-D and ARC-C reveals that models often generate correct answers but fail to adhere to a standardized format.  
Taking the Student Answer in Table~\ref{tab:llm_as_judge} as an example, the answer is semantically correct but not exactly matches the ground truth. 
This hinders automatic answer parsing and evaluation. 

To address this, we use DeepSeek-V3~\citep{deepseek-v3} as a judge to verify correctness. We validate 
this approach by manually auditing 50 random samples and observe that the LLM judgments are perfectly consistent with human evaluation. This confirms the 
judge's reliability for these tasks. For the DeepSeek-V3 content generation, we use a temperature of 0.6 and a top-p of 0.95, with a maximum response length 
of 8,192 tokens. An evaluation example is detailed in Table~\ref{tab:llm_as_judge}.

\vspace{-0.2cm}
\section{Compute Estimation}
\label{appx:flops_est}
We adopt \textbf{FL}oating-point \textbf{OP}erations (FLOPs) as our computational metric because it is hardware-agnostic and parallelization-agnostic, depending only on model architecture and sequence lengths during training.
We employ the FlopsCounter code~\citep{verl_flops_counter} of the Verl framework~\citep{verl} for estimation.
For SFT, FLOPs are estimated based on the sequence length of the SFT dataset; For RL and Syn-SFT-RL, we dynamically compute FLOPs using real-time prompt and response lengths recorded in TensorBoard~\citep{tensorboard} logs. During training, both forward and backward cost the computation.

\paragraph{Forward FLOPs Per-Token Estimation.} The the theoretical forward FLOPs per token is denoted as $\mathcal{F}_{\text{forward\_token}}$, based on the model configuration and average sequence length $S$. Let $L$ be the number of layers, $H$ the hidden size, $H_{ff}$ the intermediate size of the feed-forward network, and $V$ the vocabulary size. For the attention mechanism, we define $D_{KV}$ as the total dimension of the Key and Value heads, accounting for Grouped Query Attention (GQA)~\citep{gqa}.

First, we define the parameter counts for the constituent dense components. The MLP block, which utilizes a SwiGLU activation function with three linear projections (gate, up, and down), has a parameter count $P_{\text{MLP}}$. The linear projections in the attention layer (comprising $W_Q, W_K, W_V, W_O$) contribute $P_{\text{attn\_linear}}$. The embedding layer and the language model head share the vocabulary-dim parameters, denoted as $P_{\text{vocab}}$. These are formulated as:

\begin{small}
\begin{equation}
\begin{aligned}
    P_{\text{MLP}} &= 3 H H_{ff} \\
    P_{\text{attn\_linear}} &= H(H + 2D_{KV} + H) = 2H(H + D_{KV}) \\
    P_{\text{vocab}} &= 2 V H
\end{aligned}
\end{equation}
\end{small}

The total FLOPs consists of the dense computation part ($\mathcal{F}_{\text{dense}}$) and the attention score computation part ($\mathcal{F}_{\text{attn\_core}}$). The dense part aggregates the parameters from all $L$ layers and the vocabulary projections, multiplied by a factor of 2 (for multiply-accumulate operations). The attention core part depends linearly on the sequence length $S$. The final estimation is given by:

\begin{small}
\begin{equation}
\begin{aligned}
    \mathcal{F}_{\text{dense}} = & 2 \cdot \left[ L \cdot (P_{\text{MLP}} + P_{\text{attn\_linear}}) + P_{\text{vocab}} \right] \\
    \mathcal{F}_{\text{attn\_core}} &= 4 \cdot S \cdot L \cdot H \\
    \mathcal{F}_{\text{forward\_token}} &= \mathcal{F}_{\text{dense}} + \mathcal{F}_{\text{attn\_core}}
\end{aligned}
\end{equation}
\end{small}

\paragraph{Backward FLOPs Per-Token Estimation.}
According to~\citet{magatron_train} and~\citet{kaplan2020scalinglawsneurallanguage}, the theoretical backward FLOPs per token is approximately two times that of forward. Let $\mathcal{F}_{\text{forward\_token}}$ be the theoretical backward FLOPs per token:
\begin{equation}
    \mathcal{F}_{\text{backward\_token}} = 2\cdot \mathcal{F}_{\text{forward\_token}}
\end{equation}

\subsection{SFT Per-step Estimation}
Per-step SFT accounts for one forward and one backward pass per step. Let $B$ denote the batch size (number of sequences), $S$ the average sequence length used for fine-tuning.

The total number of tokens processed during SFT per step is given by $T_{\text{total}} = B \cdot S$. Since the backward pass requires approximately twice the FLOPs of the forward pass, the total FLOPs per token during training is $3 \cdot \mathcal{F}_{\text{forward\_token}}$.
Therefore, the total computational cost for SFT, denoted as $\mathcal{F}_{\text{SFT}}$, is calculated as:

\begin{small}
\begin{equation}
\begin{aligned}
    \mathcal{F}_{\text{train\_token}} &= \mathcal{F}_{\text{forward\_token}} + \mathcal{F}_{\text{backward\_token}} \\&= 3 \cdot \mathcal{F}_{\text{forward\_token}} \\
    \mathcal{F}_{\text{SFT}} &= B \cdot S \cdot \mathcal{F}_{\text{train\_token}} \\
               &= 3 \cdot B \cdot S \cdot \mathcal{F}_{\text{forward\_token}}
\end{aligned}
\end{equation}
\end{small}

\subsection{RL Per-step Estimation}
\paragraph{DAPO.} For DAPO, the computational cost per step is divided into a {Generation Phase} (dynamic sampling) and a {Training Phase} (actor update). 
Let $B_{\text{gen}}$ denote the generation batch size, $K$ the number of dynamic sampling iterations, and $G$ the number of responses per prompt (i.e., group size). In the generation phase, the model explores a large solution space by generating $K \cdot B_{\text{gen}} \cdot G$ sequences. Since this phase involves only inference, the cost is purely forward FLOPs.

In the training phase, a subset of data (removing all correct and wrong trajectories) is selected, denoted by the training batch size $B_{\text{train}}$ (where $B_{\text{train}} < K \cdot B_{\text{gen}}$). The update step involves one forward pass to compute new log-probs and one backward pass. Following standard estimation, the combined update cost (forward + backward) is approximately $3$ times the forward cost per token~\citep{kaplan2020scalinglawsneurallanguage}.

Given the total sequence length $S = S_{\text{prompt}} + S_{\text{response}}$, the FLOPs for one DAPO step are estimated as:

\begin{small}
\begin{equation}
\begin{aligned}
    \mathcal{F}_{\text{gen}} &= (K \cdot B_{\text{gen}} \cdot G) \cdot S \cdot \mathcal{F}_{\text{forward\_token}} \\
    \mathcal{F}_{\text{train}} &= (B_{\text{train}} \cdot G) \cdot S \cdot 3 \cdot \mathcal{F}_{\text{forward\_token}} \\
    \mathcal{F}_{\text{DAPO}} &= \mathcal{F}_{\text{gen}} + \mathcal{F}_{\text{train}} \\
     = & (K\cdot B_{\text{gen}} + 3\cdot B_{\text{train}})\cdot G \cdot S \cdot \mathcal{F}_{\text{forward\_token}} 
\end{aligned}
\end{equation}
\end{small}
\paragraph{GRPO.}
The algorithm serves as the baseline where no dynamic difficulty sampling is performed. In this setting, the generation batch size equals the training batch size ($B_{\text{gen}} = B_{\text{train}} = B$) and sampling is performed once ($K=1$). The model generates responses for all prompts in the batch and updates on all of them. Thus, the FLOPs estimation simplifies to:

\begin{small}
\begin{equation}
\label{eq:grpo_flops}
\begin{aligned}
    \mathcal{F}_{\text{gen}} &= (1 \cdot B \cdot G) \cdot S \cdot \mathcal{F}_{\text{forward\_token}} \\
    \mathcal{F}_{\text{train}} &= (B\cdot G) \cdot S \cdot 3 \cdot \mathcal{F}_{\text{forward\_token}} \\
    \mathcal{F}_{\text{GRPO}} &= \mathcal{F}_{\text{gen}} + \mathcal{F}_{\text{train}} \\
     &= 4 \cdot B \cdot G \cdot S \cdot \mathcal{F}_{\text{forward\_token}}
\end{aligned}
\end{equation}
\end{small}

\subsection{Syn-SFT-RL Per-step Estimation}
\paragraph{LUFFY and SRFT.}
Both LUFFY and SRFT integrate expert demonstrations into the RL optimization loop. Let $G$ denote the number of on-policy sampled trajectories (group size) and $N$ denote the number of expert trajectories per prompt.
In the {Generation Phase}, the model generates $G$ responses for each prompt in the batch $B$.
In the {Training Phase}, the model updates parameters using both the on-policy generated data and the off-policy expert data. Thus, the effective training batch size per prompt becomes $G + N$ ($G=7, N=1$ in~\S~\ref{appx:rl_implement}). 

Given the real-time recorded average on/off-policy sequence length $S_{\text{on}}$ and $S_{\text{off}}$, the FLOPs for LUFFY and SRFT are calculated as the sum of inference cost on $G$ samples and update cost on $G+N$ samples:

\begin{small}
\begin{equation}
\begin{aligned}
    \mathcal{F}_{\text{gen}} =& [(B \cdot G) \cdot S_{\text{on}} + (B \cdot N) \cdot S_{\text{off}} ]\cdot \mathcal{F}_{\text{forward\_token}} \\
    \mathcal{F}_{\text{train}} =& 3[(B \cdot G) \cdot S_{\text{on}} + (B \cdot N) \cdot S_{\text{off}} ] \cdot  \mathcal{F}_{\text{forward\_token}} \\
    \mathcal{F}_{\text{Hybrid}} &= \mathcal{F}_{\text{gen}} + \mathcal{F}_{\text{train}} \\
    =& 4[(B \cdot G) \cdot S_{\text{on}} + (B \cdot N) \cdot S_{\text{off}} ] \cdot \mathcal{F}_{\text{forward\_token}} 
\end{aligned}
\end{equation}
\end{small}

Note that for SRFT, although it computes multiple loss terms (Eq.~\ref{eq:srft_obj}), the dominant computational overhead remains the forward and backward passes through the transformer backbone on the combined data tokens ($G+N$), making this estimation applicable to both algorithms.

\paragraph{UPT.}
The per-step FLOPs of UPT are estimated dynamically based on the actual composition of the training batch, which consists of $N_{\text{on}}$ on-policy samples processed via GRPO and $N_{\text{off}}$ expert samples processed via SFT.

Let $S_{\text{on}}$ be the average on-policy sequence length. The algorithm processes $G\cdot S_{\text{on}}$ tokens during the Generation Phase. In this phase, the computational cost is given by:

\begin{equation}
\small
\mathcal{F}_{\text{gen}} = G \cdot S_{\text{on}} \cdot \mathcal{F}_{\text{forward\_token}}
\end{equation}
Subsequently, the algorithm filters samples based on difficulty, retaining $N{\text{on}}$ on-policy samples and $N_{\text{off}}$ off-policy samples per batch. Consequently, the FLOPs consumption during the Training Phase is:
\begin{equation}
\small
\mathcal{F}_{\text{train}} = 3 \cdot \left( N_{\text{on}}\cdot S_{\text{on}} + N_{\text{off}}\cdot S_{\text{off}} \right) \cdot \mathcal{F}_{\text{forward\_token}}
\end{equation}

Therefore, the total computational cost for a single UPT step is formulated as:

\begin{small}
\begin{equation}
\begin{gathered}\mathcal{F}_{\text{UPT}} =\mathcal{F}_{\text{gen}} +\mathcal{F}_{\text{train}}\\ =\  \left[ G\cdot S_{\text{on}}+3\cdot \left( N_{\text{on}}\cdot S_{\text{on}}+N_{\text{off}}\cdot S_{\text{off}} \right) \right] \cdot \mathcal{F}_{\text{forward\_token}}\end{gathered}
\end{equation}
\end{small}
 
\paragraph{Prefix-RFT.}
Since Prefix-RFT only borrows expert trajectory prefixes as a curriculum during RL training, 
its computational cost is otherwise identical to GRPO. We thus adopt the same FLOPs estimation 
method as in Eq.~\ref{eq:grpo_flops}.

\section{Robust Curve Fitting}
~\label{appx:robust_curve_fit}
In \S~\ref{sec:framework}, we primarily model the RL scaling dynamics using 
sigmoidal curves~\citep{ruan2024observationalscalinglawspredictability, rlscale}, 
which effectively characterize the behavior of most RL and Syn-SFT-RL runs. 
However, for unstable training instances that defy sigmoidal fitting, 
such as the SRFT case shown in Figure~\ref{fig:paradigm_compare} (Left), 
we adopt a robust fallback strategy. Specifically, we define the post-training ceiling 
for these instances as the peak performance observed either within the allocated RL compute budget or prior to performance collapse.

For the valid scaling instances, simply applying standard regression is often insufficient due to inherent training noise.
To accurately model the relationship between computational investment (FLOPs) and model performance, we employ a robust curve-fitting pipeline. 
This pipeline integrates an iterative outlier detection mechanism based on Modified Z-scores with a Least Trimmed Squares (LTS) regression optimization.

\subsection{Data Formulation}
Let $\mathcal{D} = \{(x_i, y_i)\}_{i=1}^{N}$ denote the dataset, where $x_i$ represents cumulative FLOPs and $y_i$ represents the evaluation metric. 
The data is partitioned into a training set $\mathcal{D}_{\text{train}}$ (first $N_{\text{fit}}$ points) and a held-out validation set $\mathcal{D}_{\text{val}}$. 
The train-validation split may vary due to differences in training convergence.
Across all experimental runs, the training set accounts for an average of 71\% of the data, with the remaining 29\% used for validation.

\subsection{Robust Estimation Algorithm}~\label{appx:robust_estimate}
Standard least-squares estimation is highly sensitive to anomalies. To derive a scaling law that reflects the consistent signal rather than transient noise, 
we employ a hierarchical robust optimization framework that integrates iterative statistical filtering (Modified Z-score) with subset-based optimization (Least Trimmed Squares) to isolate the true performance signal, ensuring that the derived scaling laws are predictive and generalizable across different compute regimes.
\paragraph{Stage-1: Coarse Outlier Rejection (Modified Z-Score).} 
First, we filter gross statistical anomalies. In each iteration, we compute residuals $r_i = y_i - f(x_i; \theta)$ and the median residual $\tilde{r} = \text{median}(\mathbf{r})$. To quantify deviation robustly, we calculate the Median Absolute Deviation (MAD)~\citep{huber2011robust, leys2013detecting}:

\begin{small}
    \begin{equation}~\label{sec:mad}
    \text{MAD} = \text{median}(|r_i - \tilde{r}|)
    \end{equation}
\end{small}
Subsequently, the Modified Z-score $M_i$ is computed as~\citep{iglewicz1993detect}:

\begin{small}
    \begin{equation}~\label{eq:outlier_detection}
        M_i = \frac{0.6745 \cdot (r_i - \text{median}(r))}{\text{MAD}}
    \end{equation}
\end{small}
Points with $|M_i| > \tau$ are removed from the active training split, 
where $\tau$ is the \textbf{outlier threshold}.
The factor $0.6745$ normalizes the score such that it is consistent with 
the standard deviation under a normal distribution, while the use of MAD ensures resilience against extreme values that would skew a standard variance calculation.

\paragraph{Stage-2: Least Trimmed Squares Regression.}
To further refine the model against subsets of data that may distort the global trend, we employ Least Trimmed Squares (LTS)~\citep{rousseeuw1984least, rousseeuw1987robust} regression. Instead of minimizing the sum of all residuals, LTS regression minimizes only the smallest $h$ squared residuals:

\begin{small}
    \begin{equation}~\label{eq:lts_obj}
    \hat{\theta}_{\text{LTS}} = \underset{\theta}{\arg\min} \sum_{j=1}^{h} (r^2)_{(j)}(\theta)
    \end{equation}
\end{small}
\noindent where $(r^2)_{(1)} \le \dots \le (r^2)_{(N_{\text{fit}})}$ are the ordered squared residuals over the training set, and $h = \lfloor N_{\text{fit}} \cdot \alpha \rfloor$ is 
determined by the parameter $\mathbf{\alpha}$. We define ${H^{(k+1)}}$ as the set of indices corresponding to the $h$ smallest squared residuals, i.e., $H^{(k+1)} = \{i \mid r_i^2 \le (r^2)_{(h)}\}$.  

We optimize this objective using the Concentration Step (C-step) algorithm \citep{rousseeuw2006computing}, which proceeds iteratively as follows:

\begin{itemize}[leftmargin=12pt, nosep]
    \item \textbf{Estimation}: Compute squared residuals $r_i^2 = (y_i - f(x_i; \theta^{(k)}))^2$ for all $N_{\text{fit}}$ training points using the current parameters $\theta^{(k)}$.
    \item \textbf{Selection}: Identify the index set $H^{(k+1)}$ corresponding to the $h$ smallest squared residuals.
    \item \textbf{Update}: Update parameters to $\theta^{(k+1)}$ by fitting the model strictly to the data points indexed by $H^{(k+1)}$.
    \item \textbf{Convergence}: Repeat the process until the parameter estimate $\theta$ stabilizes.
\end{itemize}
\FloatBarrier

\newpage
\begin{table*}[!htbp]
\centering 
\caption{\textbf{Scaling dynamics across SFT-then-RL configurations.} 
The first row represents the Pure-RL baseline. $x_{\text{sft}}$ denotes cumulative SFT compute (exaFLOPs). 
\textbf{Outlier Threshold} is the Modified Z-score cutoff $\tau$ for Stage-1 outlier rejection (points with $|M_i|>\tau$ are removed; Appx.~\ref{appx:robust_estimate}). 
Use-LTS denotes whether the Least Trimmed Squares (LTS) regression in Appx.~\ref{appx:robust_estimate} is used during curve fitting. 
In \S~\ref{sec:dynamics} we define the alignment offset $\epsilon$, $PL_{\text{rl}}$, $A_{\text{post}}$, and the midpoint parameter $C_{\text{mid}_{\text{rl}}}$ in the RL scaling law (Eq.~\ref{eq:rl_scaling}). 
For $PL_{\text{rl}}$, $B_{\text{rl}}$, $P_{\text{sft}}$, and $A_{\text{post}}$, higher values are better; for $C_{\text{mid}_{\text{rl}}}$, lower values are better. 
The best values within each SFT configuration are bolded.}
\label{tab:fit_res_tab}
\resizebox{\textwidth}{!}{
    \begin{tabular}{@{}ccccccccccc@{}}
        \toprule
        SFT data                     & SFT Step & $x_{\text{sft}}$ & Outlier Threshold & Use-LTS               & $\epsilon$ & $PL_{\text{rl}}$ & $C_{\text{mid}_{\text{rl}}}$ & $B_{\text{rl}}$           & $P_{\text{sft}}$ & $A_{\text{post}}$ \\ \midrule
        -                            & 0        & 0                                       & 2.5               & False                 & 0          & 25.3                                        & 1                & 1.2           & 46.7             & 77.3                                   \\ \midrule
        \multirow{5}{*}{S1K}         & 62       & 0.6                                     & 2.5               & False                 & 4.4        & \textbf{8.7}                                & 107              & 8.5           & 68.6             & \textbf{77.0}                          \\
                                     & 124      & 1.1                                     & 3                 & True ($\alpha=$0.85)  & 0          & 6.0                                         & \textbf{44}      & 0.9           & 70.9             & 76.9                                   \\
                                     & 186      & 1.7                                     & 2.5               & False                 & 0          & 4.0                                         & 99               & 1.1           & 72.9             & 76.5                                   \\
                                     & 248      & 2.3                                     & 2.5               & True  ($\alpha=$0.75) & 0          & 2.5                                         & 93               & 2.8           & \textbf{74.1}    & 76.4                                   \\
                                     & 310      & 2.9                                     & 3                 & True ($\alpha=$0.75)  & 1.5        & 3.3                                         & 134              & \textbf{11.2} & 73.1             & 75.9                                   \\ \midrule
        \multirow{5}{*}{Easy102K}    & 360      & 19.8                                    & 2.5               & False                 & 0          & \textbf{6.0}                                & \textbf{22}      & 1.0           & 70.7             & 76.7                                   \\
                                     & 720      & 39.5                                    & 2.5               & False                 & 0          & 5.3                                         & 66               & 0.8           & 73.3             & 78.5                                   \\
                                     & 1080     & 59.3                                    & 2.5               & False                 & 0          & 4.6                                         & 135              & 0.4           & 74.2             & \textbf{78.8}                          \\
                                     & 1440     & 79.0                                    & 2.5               & True ($\alpha=$0.75)  & 0          & 2.0                                         & 41               & \textbf{2.0}  & \textbf{74.8}    & 76.8                                   \\
                                     & 1800     & 98.8                                    & 2.5               & False                 & 0          & 3.4                                         & 43               & 0.3           & 74.4             & 77.8                                   \\ \midrule
        \multirow{5}{*}{Uniform102K} & 360      & 34.7                                    & 2.5               & False                 & 0          & \textbf{4.8}                                & 32               & 1.8           & 72.2             & 76.9                                   \\
                                     & 720      & 69.3                                    & 3                 & True ($\alpha=$0.75)  & 0          & 3.0                                         & \textbf{28}      & 1.4           & 74.5             & 77.6                                   \\
                                     & 1080     & 104.0                                   & 2.5               & False                 & 1.5        & 4.8                                         & 118              & \textbf{6.6}  & 74.8             & \textbf{79.7}                          \\
                                     & 1440     & 138.7                                   & 2.5               & False                 & 0          & 1.8                                         & 46               & 2.2           & \textbf{75.7}    & 77.5                                   \\
                                     & 1800     & 173.4                                   & 2.5               & True ($\alpha=$0.75)  & 0          & 3.8                                         & 161              & 0.7           & 75.5             & 79.3                                   \\ \midrule
        \multirow{5}{*}{Hard102K}    & 360      & 89.3                                    & 4                 & False                 & 0          & 4.9                                         & 40               & 0.9           & 73.2             & 78.1                                   \\
                                     & 720      & 178.5                                   & 4                 & True ($\alpha=$0.85)  & 0          & 3.1                                         & \textbf{32}      & 1.7           & 74.7             & 77.8                                   \\
                                     & 1080     & 267.8                                   & 3                 & True ($\alpha=$0.85)  & 0          & \textbf{6.7}                                & 61               & 3.5           & 76.2             & \textbf{82.8}                          \\
                                     & 1440     & 357.0                                   & 2.5               & True ($\alpha=$0.85)  & 0          & 3.4                                         & 48               & \textbf{5.5}  & \textbf{76.9}    & 80.3                                   \\
                                     & 1800     & 446.3                                   & 2.5               & False                 & 0          & 2.7                                         & 73               & 1.0           & 76.1             & 78.7                                   \\ \midrule
        \multirow{12}{*}{SFT889K}    & 360      & 34.9                                    & 2.5               & False                 & 0          & \textbf{13.4}                               & 10               & 0.7           & 69.2             & 82.5                                   \\
                                     & 720      & 69.8                                    & 3                 & True ($\alpha=$0.75)  & 0          & 8.8                                         & \textbf{8}       & 1.0           & 71.9             & 80.7                                   \\
                                     & 1080     & 104.8                                   & 4                 & True ($\alpha=$0.75)  & 0          & 7.1                                         & 10               & 1.5           & 73.0             & 80.1                                   \\
                                     & 1440     & 139.7                                   & 4                 & True ($\alpha=$0.75)  & 0          & 7.0                                         & 9                & 1.7           & 73.3             & 80.3                                   \\
                                     & 1800     & 174.6                                   & 3                 & True ($\alpha=$0.75)  & 0          & 6.9                                         & 9                & 1.6           & 74.1             & 81.0                                   \\
                                     & 3600     & 349.2                                   & 4                 & False                 & 0          & 6.6                                         & 13               & 1.6           & 76.5             & 83.1                                   \\
                                     & 5400     & 523.8                                   & 2.5               & False                 & 0          & 5.3                                         & 13               & 2.2           & 77.5             & 82.8                                   \\
                                     & 7200     & 698.4                                   & 2.5               & True ($\alpha=$0.75)  & 0          & 6.7                                         & 16               & 1.6           & 77.3             & \textbf{84.0}                          \\
                                     & 9000     & 873.0                                   & 2.5               & False                 & 0          & 6.7                                         & 13               & 1.8           & 76.7             & 83.4                                   \\
                                     & 10800    & 1047.6                                  & 2.5               & True ($\alpha=$0.85)  & 0          & 5.4                                         & 12               & \textbf{2.5}  & \textbf{77.7}    & 83.1                                   \\
                                     & 12600    & 1222.2                                  & 3                 & True ($\alpha=$0.75)  & 0          & 5.9                                         & 10               & 2.2           & 77.1             & 83.0                                   \\
                                     & 14080    & 1365.8                                  & 2.5               & True ($\alpha=$0.75)  & 0          & 7.8                                         & 11               & 1.1           & 76.2             & 84.0                                   \\ \bottomrule
    \end{tabular}
}
\end{table*}

\subsection{Fitting Results and Analysis.}
\subsubsection{Fitting Results}
Across all fitting instances, the pipeline achieves an average Root Mean Square Error (RMSE) of 
approximately 0.86 on the validation split and an average $R^{2}$ of 0.90 on the training split, 
confirming the overall robustness of the fits. 

\subsubsection{Fitting Analysis}
\begin{figure*}[t!] 
    \centering
    \begin{subfigure}[b]{0.48\textwidth}
        \centering
        \includegraphics[width=\linewidth]{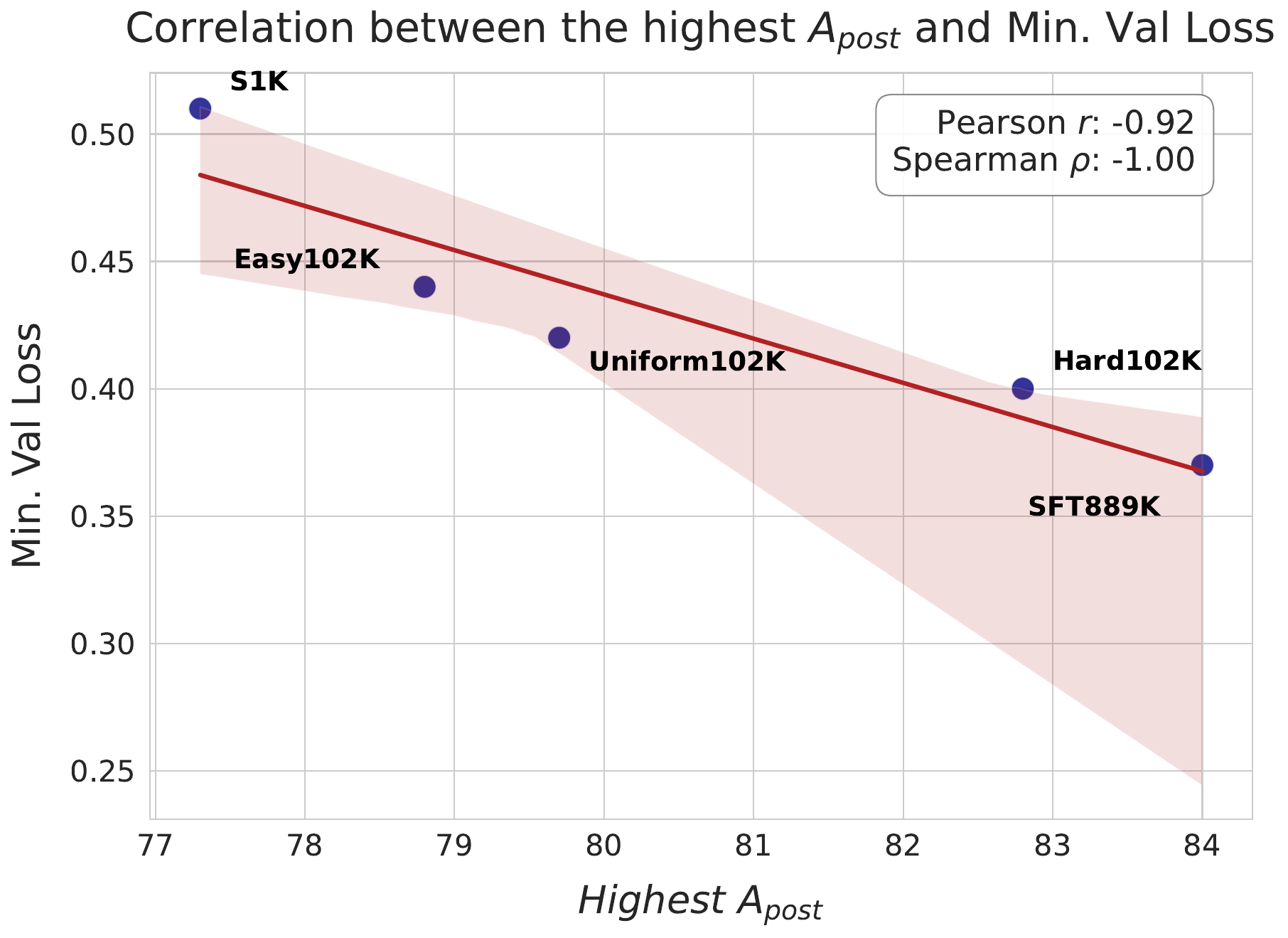}
        \caption{Correlation Analysis}
        \label{fig:correlation}
    \end{subfigure}
    \hfill 
    \begin{subfigure}[b]{0.48\textwidth}
        \centering
        \includegraphics[width=\linewidth]{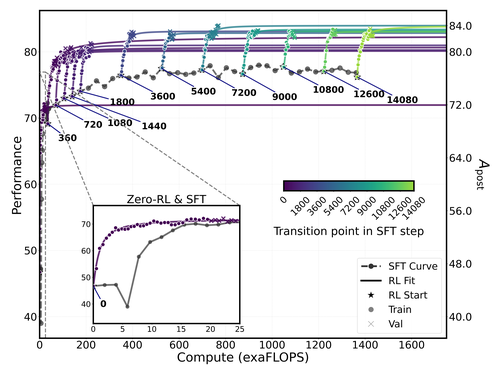}
        \caption{SFT889K}
        \label{fig:general}
    \end{subfigure}
    
    \vspace{10pt} 
    
    \begin{subfigure}[b]{0.48\textwidth}
        \centering
        \includegraphics[width=\linewidth]{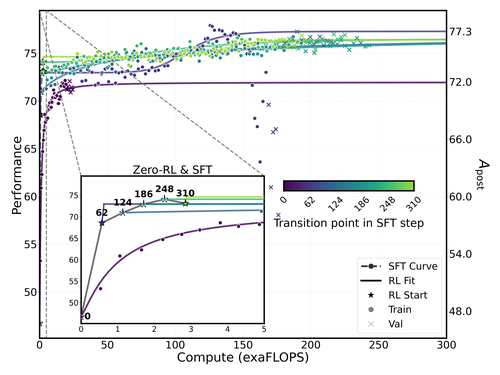}
        \caption{S1K}
        \label{fig:s1k}
    \end{subfigure}
    \hfill
    \begin{subfigure}[b]{0.48\textwidth}
        \centering
        \includegraphics[width=\linewidth]{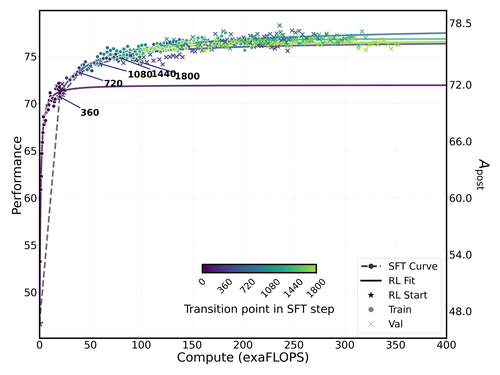}
        \caption{Easy102K}
        \label{fig:easy}
    \end{subfigure}
    \vspace{10pt} 
    \begin{subfigure}[b]{0.48\textwidth}
        \centering
        \includegraphics[width=\linewidth]{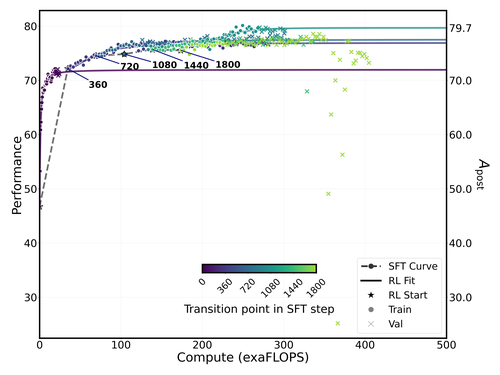}
        \caption{Uniform102K}
        \label{fig:uniform}
    \end{subfigure}
    \hfill
    \begin{subfigure}[b]{0.48\textwidth}
        \centering
        \includegraphics[width=\linewidth]{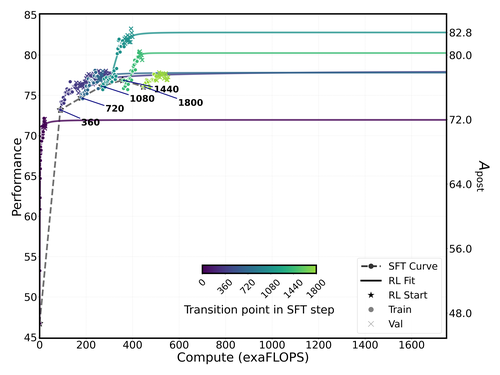}
        \caption{Hard102K}
        \label{fig:hard}
    \end{subfigure}
    \caption{\textbf{Visualization of SFT-then-RL fitting on Qwen2.5-7B across different SFT data configurations.} 
    (\textbf{a}) Correlation analysis between the highest $A_{\text{post}}$ and the minimum validation loss of each specified SFT dataset. 
    (\textbf{b})-(\textbf{f}) The SFT-then-RL scaling dynamics under various data configurations. The SFT trajectory is depicted by a black dashed line. 
    RL scaling curves initiated from different SFT steps are distinguished by a color gradient, where lighter shades indicate a higher number of 
    SFT steps. The specific starting SFT step count for each RL curve is annotated in bold black text. Data points from the training split used for 
    fitting the RL scaling curves are marked with solid circles, while those from the validation split used for assessing curve goodness-of-fit are marked 
    with crosses. Magnified views are provided for the low-compute regions of SFT889K and S1K.}
    \label{fig:fitting}
\end{figure*}

We visualize the resulting SFT-then-RL
scaling curves in Figure~\ref{fig:fitting} and provide the exhaustive parameter estimates 
in Table~\ref{tab:fit_res_tab}. This table systematically reports the evolution of key scaling 
metrics across various SFT data configurations relative to the progress of SFT compute $x_{\text{sft}}$. 
Specifically, we characterize the realized SFT performance ($P_{\text{sft}}$), the RL plasticity ($PL_{\text{rl}}$), 
the RL convergence midpoint ($C_{\text{mid}_{\text{rl}}}$), 
the RL scaling steepness ($B_{\text{rl}}$), and the ultimate post-training ceiling ($A_{\text{post}}$). 
These results facilitate a detailed analysis of how the SFTed initial point impacts 
the subsequent RL trend and the final performance limits of the model.

\paragraph{Impact of SFT Compute on RL Plasticity.}
As illustrated in the $PL_{\text{rl}}$ column of Table~\ref{tab:fit_res_tab}, RL plasticity exhibits an overall downward trend, 
albeit with minor oscillations, as SFT compute $x_{\text{sft}}$ increases. Consequently, allocating minimal compute to SFT generally 
preserves greater plasticity and yields a larger headroom for subsequent RL-driven improvements. Specifically, across the S1K, Easy102K, 
Uniform102K, and SFT889K configurations, the lowest SFT compute consistently achieves the maximum $PL_{\text{rl}}$. Although Hard102K 
presents a minor deviation, its earliest SFT checkpoint still retains the second-highest plasticity in that category. 

We attribute this 
general decline to the inherent imitation nature of SFT. The SFT progress deeply entrenches the policy within the demonstration 
distribution, prematurely exhausting the exploratory capacity required for effective reward-driven optimization~\citep{wang2026learningstayingcuriousentropypreserving}. This hypothesis is 
empirically corroborated by the Pure-RL baseline. By entirely bypassing the SFT phase, the Pure-RL instance retains maximum exploratory 
freedom and achieves the highest absolute plasticity of 25.3, outperforming all SFT-then-RL configurations. However, this plasticity 
advantage is ultimately offset by its low initial performance ($P_{\text{sft}} = 46.7$), which severely restricts the final post-training 
ceiling ($A_{\text{post}}$).

\paragraph{Impact of SFT Compute on RL Training Efficiency.}
We quantify RL training efficiency using the convergence midpoint $C_{\text{mid}_{\text{rl}}}$, 
which measures the amount of compute required to achieve half of the total potential gain. 
Lower values of $C_{\text{mid}_{\text{rl}}}$ indicate faster convergence.
However, $C_{\text{mid}_{\text{rl}}}$ does not monotonically decrease with increasing $x_{\text{sft}}$, 
instead exhibiting substantial fluctuations.

As shown in Table~\ref{tab:fit_res_tab}, the most efficient RL runs, those with the 
lowest $C_{\text{mid}_{\text{rl}}}$, typically correspond to relative less allocated SFT compute $x_{\text{sft}}$.
Notably, the Pure-RL baseline achieves the lowest $C_{\text{mid}_{\text{rl}}}$ of 1 exaFLOPs, 
representing the fastest convergence among all evaluated scenarios. 
This phenomenon is broadly consistent with prior observations regarding the 
interplay between SFT and subsequent RLHF training~\citep{NEURIPS2022_b1efde53}. 
It may be attributed to the fact that relatively lower SFT compute 
preserves more diverse behaviors in the action space, thereby facilitating exploration and 
accelerating RL convergence.

\section{Dataset Curation}
\label{dataset_summary}
We summarize the key characteristics of our SFT data in Table~\ref{tab:data_character}.

\subsection{Expert Trajectory Collection}
\label{appx:sft_data_curate}
We curate high-quality reasoning trajectories from two large-scale datasets: AM-DeepSeek-R1-Distilled-1.4M~\citep{amthink1_4m} and AM-DeepSeek-Distilled-40M~\citep{amthink40m}.
To ensure data distribution consistency and quality, we retain only mathematics-domain data, select trajectories distilled from DeepSeek-R1-671B~\citep{deepseek-r1} to unify trajectory style, and perform deduplication based on prompt matching.
The resulting filtered datasets are denoted as amthink-1.4m and amthink-40m.
From these sources, we construct multiple datasets for our experiments: SFT889K, Uniform102K, Easy102K, and Hard102K for SFT training, and Val-199 for SFT validation.
A prompt-trajectory example is shown in Table~\ref{tab:sft_data_exp}.

\paragraph{Difficulty Classification.}
To understand the influence of SFT data difficulty on post-training outcomes, we extract data of varying difficulty levels from amthink-40m. 
We use the Win Rate (WR), defined as the ratio of successful attempts $S$ to total attempts $N$, as a proxy for problem difficulty. 
We derive WRs with $N$=4 attempts across three models: DeepSeek-R1-Distill-1.5B, 7B, and DeepSeek-R1-671B~\citep{deepseek-r1}. 
Based on WRs from the 1.5B model, problems are classified as Easy ($\mathrm{WR}$=1.0) or Hard ($\mathrm{WR}\le$ 0.25). 
Consequently, we construct Easy102K and Hard102K, each containing 102K samples from their respective pools, alongside Uniform102K, a randomly sampled neutral-difficulty baseline.

To verify that our WR-based classification captures genuine reasoning complexity rather than mere obscurity, we conduct a supplementary semantic analysis. 
We randomly sample 10,000 prompts from both Easy102K and Hard102K, employing DeepSeek-V3.2~\citep{deepseekv3.2} as an expert evaluator to decompose solution trajectories into distinct \textbf{logical steps} while filtering out conversational redundancy. 
The evaluator follows the fixed instruction prompt in Table~\ref{tab:logical_steps_prompt}.
Manual verification of 100 sampled outputs per dataset confirms that this automated step-counting aligns closely with human judgment. 

As shown in Table~\ref{tab:sft_data_exp}, trajectories in Hard102K contain an average of 13 logical steps, representing a 30\% increase over the 10 steps averaged in Easy102K. 
This reasoning depth is directly mirrored in trajectory length: Hard102K responses average 8,532 tokens, nearly four times the 2,253 tokens typical of Easy102K. 
This concurrent increase in both logical steps and trajectory length empirically validates WR as a robust and reliable proxy for problem difficulty in our study.

\subsection{RL Data}
We curate RL62K (62.3K prompts) from Skywork-OR1-RL~\citep{skywork-or1} by filtering out extreme difficulty levels and prompts containing Chinese characters.
For {Syn-SFT-RL}, we construct MIX37K (36.7K samples) by augmenting these prompts with matched expert trajectories from SFT889K, excluding sequences exceeding 8{,}192 tokens as suggested by~\citep{yan2025learningreasonoffpolicyguidance} to ensure the complete trajectory utilization in each update step.
Crucially, this data scale is sufficient, as our experiments show that RL variants typically reach saturation or instability before exhausting a single epoch over MIX37K.

\section{Complemetary Results}
~\label{appx:complementary_results}
\vspace{-1cm}
\subsection{Paradigm Comparsion on SFT889K}
~\label{appx:paradigm_compare_sft889k}
We compare various expert trajectory utiliztion paradigms by training on SFT889K. 
In this setting, we set the 
batch size to 64 and maximum response length to 12,000, with RL capped at 
400 steps (see Appx.~\ref{appx:train_config} for details).

\begin{figure*}[t] 
    \centering
    \includegraphics[width=0.7\linewidth]{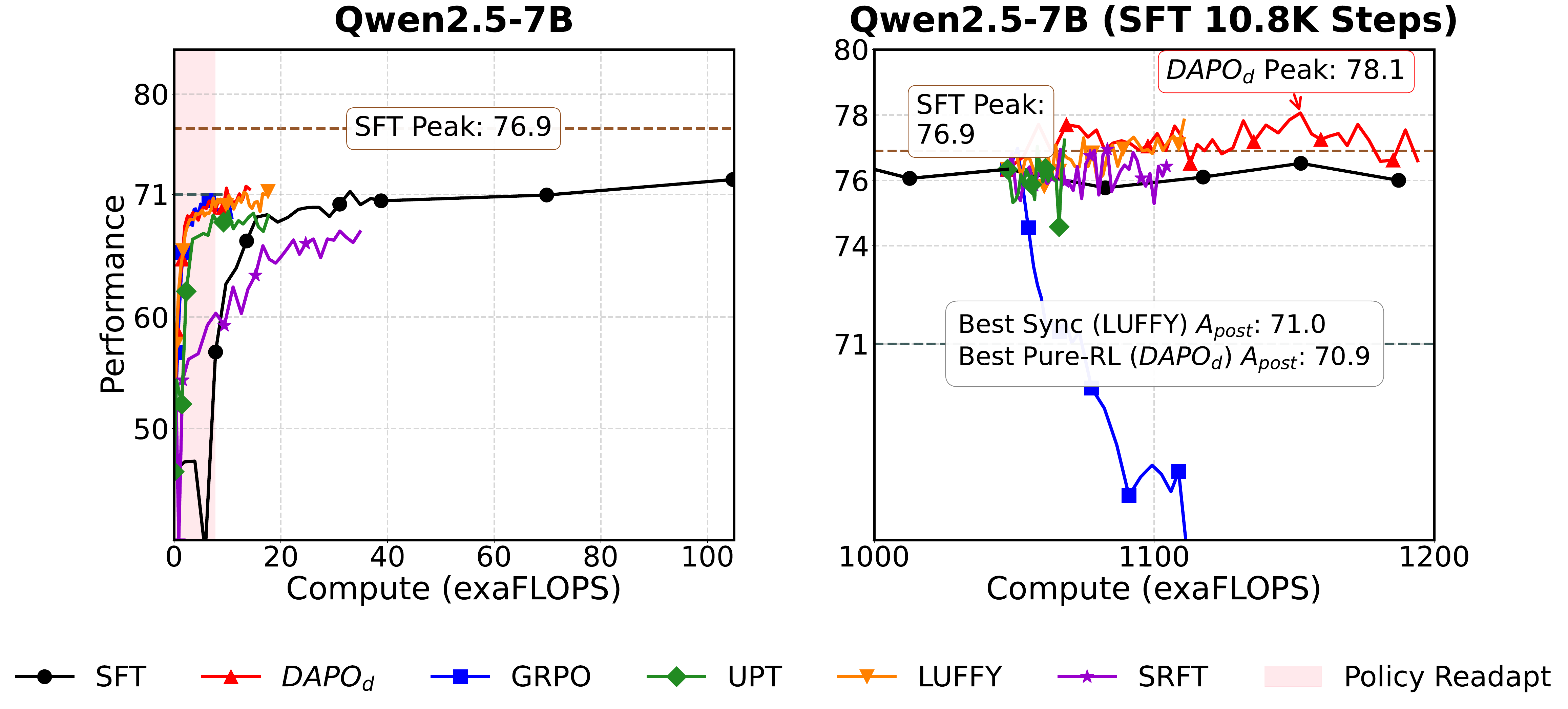}
    \caption{\textbf{The training dynamic comparision of post-training paradigms on the SFT889K dataset, covering different initializations.} 
    (\textbf{Left}) Initializing from Qwen2.5-7B, where the all RL and Syn-SFT-RL methods suffer from the prematurely convergence. 
    (\textbf{Right}) Initializing from a saturated SFT checkpoint (10,800 steps on Qwen2.5-7B), where DAPO$_{d}$ outperforms other methods, and 
    GRPO suffers from the performance collapse.}
    \label{fig:paradigm_compare_large}
\end{figure*}

As shown in Figure~\ref{fig:paradigm_compare_large} (Left), when starting from the base model, 
both RL and Syn-SFT-RL methods exhibit premature convergence within 40 exaFLOPs. $\text{DAPO}_d$ and 
LUFFY reach a ceiling ($A_{\text{post}}$) of 71.0 approximately, slightly lower than their MIX37K counterparts 
(74.3 and 72.7 in Figure~\ref{fig:paradigm_compare}). However, UPT shows significant scaling gains on this 
larger dataset (68.6 vs. 60.0), and SRFT demonstrates improved stability compared to MIX37K.

When initialized from a saturated SFT checkpoint (Figure~\ref{fig:paradigm_compare_large}, Right), GRPO 
collapses rapidly, whereas $\text{DAPO}_d$ scales steadily to 78.1. This not only outperforms the SFT peak 
(76.9) by 1.2 points but also surpasses all other baselines. These results highlight the necessity of $\text{DAPO}_d$'s 
dynamic difficulty sampling for stability and emphasize the SFT-then-RL pipeline as the superior paradigm 
for pushing post-training frontiers.

\subsection{SFT Data Scale Impact}
\label{appx:further_data_scale_study}

\begin{figure*}[t]
    \centering
    \includegraphics[width=0.7\linewidth]{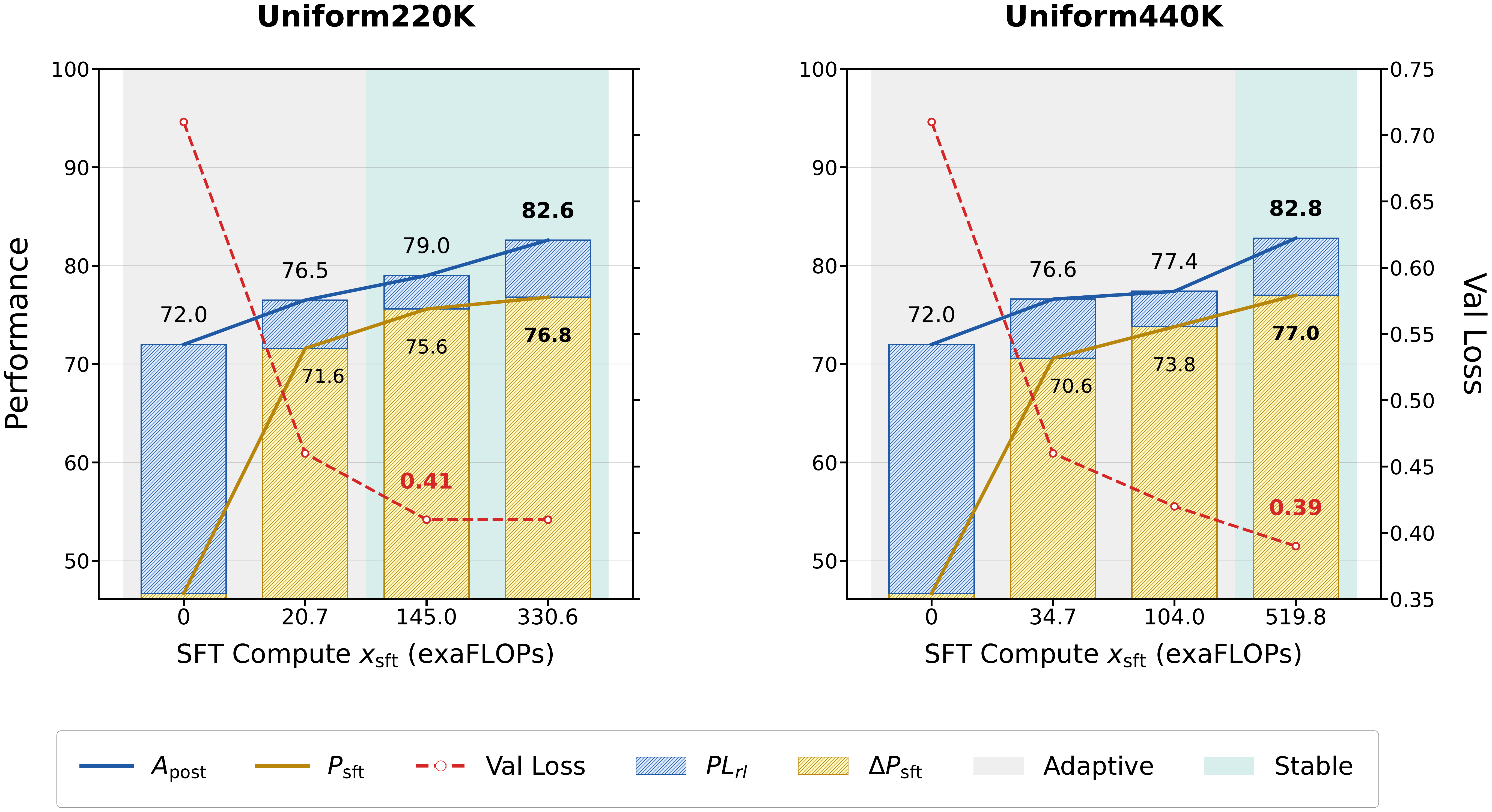}
    \caption{\textbf{SFT compute scaling dynamics of the SFT-then-RL pipeline across different scale of the SFT data.} 
            (\textbf{Left}) Performaning SFT on Uniform220K, selecting the 215-step (20.7 exaFLOPs), 1505-step (145.0 exaFLOPs) and 3432-step (330.6 exaFLOPs) SFTed checkpoints to initialize the RL training.
            (\textbf{Right}) Performing SFT on Uniform440K, selecting the 360-step (34.7 exaFLOPs), 1080-step (104.0 exaFLOPs) and 5400-step (519.8 exaFLOPs) SFTed checkpoints to initialize the RL training.
            }
    \label{fig:uniform_vary_datasets}
\end{figure*}

In \S~\ref{sec:data_impact}, we observe a distinct stratification in the SFT-then-RL post-training performance ceilings across three 
data scales: S1K, Uniform102K, and SFT889K, suggesting that a larger scale begets a higher post-training ceiling. 

To further investigate how the SFT 
volume dictates the ceiling, we introduce Uniform220K and Uniform440K, SFT datasets comprising 220K and 440K prompt-trajectory pairs uniformly drawn 
from amthink-40m. We maintain hyperparameter consistency with the Uniform102K setup (see \S~\ref{appx:sft_config}), and we save 
checkpoints every 215 steps (0.5 epochs) for Uniform220K and 360 steps (0.4 epochs) for Uniform440K. Using validation loss as a guide, we select three 
checkpoints across distinct SFT regime for each scale to initialize the subsequent RL stage.

\begin{table}[!htbp]
\centering
\small
\caption{\textbf{The relation between the highest $A_\text{post}$ and the minimum 
validation loss across different SFT data scales.} Increasing the training data scale of SFT leads 
to a lower minimal validation loss, which indicates a higher post-training ceiling $A_\text{post}$.}
\label{tab:data_scale_Apost_mvl_relation}
\begin{tabular}{@{}ccc@{}}
\toprule
SFT data    & Highest $A_\text{post}$ & Min. Val Loss \\ \midrule
S1K         & 77.3            & 0.51          \\
Uniform102K & 79.7            & 0.42          \\
Uniform220K & 82.6            & 0.41          \\
Uniform440K & 82.8            & 0.39          \\
SFT889K     & \textbf{84.0}   & \textbf{0.37} \\ \bottomrule
\end{tabular}
\end{table}

As Figure~\ref{fig:uniform_vary_datasets} illustrates, initializing RL from the Stable Regime consistently yields the optimal performance ceiling 
on both datasets. This success stems from a synergy between superior SFT performance and preserved RL plasticity, validating the significance of 
\textbf{training SFT to saturation}. As shown in Table~\ref{tab:data_scale_Apost_mvl_relation}, the highest $A_{\text{post}}$ continues to trend upward 
as data scale increases, confirming that the \textbf{larger scale, higher ceiling} principle remains robust. Furthermore, we confirm that \textbf{the minimum validation 
loss still serves as a robust indicator for pursuing a higher $A_{\text{post}}$}. Even as the SFT data scale expands, the minimum validation loss remains strongly correlated with the 
highest $A_{\text{post}}$ of each specified SFT dataset, maintaining a Pearson correlation coefficient of -0.94, and a Spearman correlation coefficient of -1.00.

\subsection{The Optimal SFT-to-RL Transition Timing on Larger Model}
\label{appx:larger_model_saturation}
\begin{table*}[!htbp]
\centering
\small
\caption{\textbf{Qwen2.5-14B on Uniform55K:} SFT-to-RL transition points. Columns report cumulative SFT compute $x_{\text{sft}}$ (exaFLOPs), validation loss, SFT regime, realized SFT performance ($P_{\text{sft}}$), maximum RL improvement over the SFT checkpoint, and peak post-training performance ($\max P_{\text{post}}$). Boldface marks the strongest outcome within each column where a single optimum is intended.}
\label{tab:qwen14b_res}
\begin{tabular}{@{}lcccccc@{}}
\toprule
SFT--RL switch & $x_{\text{sft}}$ & Val loss & SFT regime & $P_{\text{sft}}$ & Max RL improve. & $\max P_{\text{post}}$ \\ \midrule
0-step (zero-RL) & 0 & 0.67 & Adaptive & 59.8 & \textbf{14.7} & 74.5 \\
172-step & 32.7 & 0.44 & Adaptive & 75.7 & 3.1 & 78.8 \\
1720-step & 326.6 & \textbf{0.39} & Stable & \textbf{78.6} & 1.2 & \textbf{79.8} \\ \bottomrule
\end{tabular}
\end{table*}

In this section, we validate whether the principle of training the SFT phase to saturation still holds for larger models.
We construct a training set of 55K uniformly sampled instances from the amthink-40m dataset, denoted as Uniform55K and 
perform SFT-then-RL experiments on Qwen2.5-14B~\citep{qwen2025qwen25technicalreport}. 
We follow the SFT-then-RL configuration used for Qwen2.5-7B + SFT889K 
(detailed in Appx.~\ref{appx:sft_config} and \ref{appx:rl_implement}), and fine-tune the Qwen2.5-14B for four 
epochs. To evaluate the transition timing, we selected SFT checkpoints from various convergence 
regimes to initialize the subsequent RL phase. The subsequent RL is trained for 200 steps.

The established SFT foundation ($P_{\text{sft}}$) and the peak post-training performance ($\max P_{\text{post}}$) 
are summarized in Table~\ref{tab:qwen14b_res}. Consistent with our observations on Qwen2.5-7B in Figure~\ref{fig:vary_datasets} and Llama3.2-3B in Figure~\ref{fig:llama_sft_then_rl}, 
the principle of training the SFT foundation to saturation holds firmly at the 14B scale. 
We observe that transitioning to RL from the 1720-step checkpoint, which is located in the stable regime, 
achieve the best overall performance ($\max P_{\text{post}}$= 79.8). 
This optimal ceiling is primarily attributed to the robust realized SFT performance 
of 78.6. In contrast, while initiating RL from a zero-shot foundation results in the 
highest absolute maximum RL improvement of 14.7 points, its weak initial foundation 
of 59.8 prevented it from reaching a comparable end-to-end performance.

\newpage

\newpage
\begin{table*}[t]
\centering
\small
\begin{tabularx}{\linewidth}{>{\RaggedRight\arraybackslash}X}
\toprule
\textbf{System Prompt:}\\
You are a teacher grading student assignments. I will provide you with the question, the student's answer, and the ground truth for the question. 
Please carefully review each question and grade the student's answer based on the ground truth. If the student's answer matches the ground truth, reply with True; otherwise, reply with False. You need not provide any additional information---only reply with True or False. Your grading response must strictly follow the format: 'Grade: '.\\
{Example:}\\
Question: 1+1 = $\_$. Please choose the correct option from the following options. A.2, B.3, C.4\\
Student answer: I think it is A.2\\
Ground truth: A\\
Grade: True\\
\textbf{User Prompt:}\\
Question:\\
Which change is the best example of a physical change?\\
1) a cookie baking 2) paper burning 3) ice cream melting 4) a nail rusting\\
\textbf{Student answer:}\\
Okay, so I need to figure out which of these options is a physical change. Let me start by recalling what a physical change is. From what I learned, a physical change is when the form or state of a substance changes, but the chemical composition remains the same. That means no new substances are formed. In contrast, a chemical change involves a chemical reaction where new substances are created.\\
Now let's look at each option one by one.\\
1) A cookie baking. Hmm, when you bake a cookie, the dough is transformed. The ingredients like flour, eggs, sugar, etc., are mixed and then heated. Baking usually involves chemical reactions, like the Maillard reaction, where new compounds are formed. So this is probably a chemical change.\\
2) Paper burning. Burning paper is definitely a chemical change. Fire requires fuel and oxygen, and when paper burns, it combines with oxygen to form ash, carbon dioxide, and other products. The original paper is no longer there; it's changed into different substances. So this is chemical.\\
3) Ice cream melting. Ice cream is frozen, and when it melts, it goes from solid to liquid. No chemical reactions are happening here. The molecules are just spreading out because the temperature increased, but the actual molecules of the ice cream (like water, fat, sugar) are the same. So this should be a physical change.\\
4) A nail rusting. Rusting is a chemical process. The iron in the nail reacts with oxygen and water in the air to form iron oxide. This creates a new substance, which is rust. So that's a chemical change.\\
So the answer should be option 3, ice cream melting. It's a physical change because the state changes without forming new chemicals.\\
{Answer:} 3) ice cream melting\\
{Ground truth:} 3\\
{Grade:}\\
\textbf{DeepSeek-V3 Response:}\\
True\\
\bottomrule
\end{tabularx}
\caption{\textbf{The example of using DeepSeek-V3 as a judge when evaluating the case of ARC-C.} The student answer is given by the 10800-step SFTed checkpoint from Qwen-2.5-7B.}
\label{tab:llm_as_judge}
\end{table*}

\newpage
\begin{table*}[t]
\centering
\small
\begin{tabularx}{\linewidth}{>{\RaggedRight\arraybackslash}X}
\toprule
\textbf{Question:}\\
Let the set $A=\{1,2,3\}$, and the set $B=\{0,1,2\}$. Then, the number of elements in $A \cup B$ is:\\
A: 2; B: 3; C: 4; D: 5 \\
\textbf{Expert Trajectory:}\\
<think>
Okay, so I have this problem here about sets A and B. Let me read it again to make sure I understand. It says, ``Let the set A be $\{1, 2, 3\}$, and the set B be $\{0, 1, 2\}$. Then, the number of elements in $A \cup B$ is ( ). The options are A: 2, B: 3, C: 4, D: 5.'' Hmm, union of A and B. Right, union means all the elements that are in either set A or set B or in both. So I need to combine them and remove any duplicates. Let me write down the sets first. Set A is $\{1, 2, 3\}$. So those are the numbers 1, 2, and 3. Set B is $\{0, 1, 2\}$. That has 0, 1, and 2. Now, if I combine these two sets, I should list all the elements from both, but make sure I don't count the same element more than once. Let me list them out. Starting with set A: 1, 2, 3. Then set B has 0, 1, 2. So adding those to the union, we already have 1, 2, 3 from A, and from B we have 0 which isn't in A. The 1 and 2 from B are already in A, so they don't need to be added again. So combining them, the union should have elements 0, 1, 2, 3. That's four elements. Wait, but let me check again. Let me list them all out: $A \cup B = \{0, 1, 2, 3\}$. So the elements are 0, 1, 2, 3. Let's count them: 0 is one, 1 is two, 2 is three, 3 is four. So that's four elements. So the number of elements is 4. The options are C: 4. So the answer should be C? Wait, but let me make sure I didn't miss anything. Let me recall the formula for the number of elements in a union: $|A \cup B| = |A| + |B| - |A \cap B|$. So the size of the union is the size of A plus the size of B minusthe size of their intersection. Let's apply that. Set A has 3 elements: 1, 2, 3. Set B has 3 elements: 0, 1, 2. The intersection of A and B would be the elements common to both, which are 1 and 2. So $|A \cap B|$ is 2. Then $|A \cup B| = 3 + 3 - 2 = 4$. Yes, that matches what I got before. So that's confirming it. Therefore, the number of elements in $A \cup B$ is 4. So the answer is C. But wait, just to double-check, maybe I made a mistake in the intersection. Let me verify the intersection. $A \cap B$ is the set of elements that are in both A and B. A is $\{1,2,3\}$, B is $\{0,1,2\}$. So elements common to both are 1 and 2. So yes, two elements. So the formula gives $3 + 3 - 2 = 4$. That seems correct. And listing them out gives four elements. So I think that's right. Alternatively, maybe the answer is D: 5? Wait, maybe I miscounted. Let me list the elements again. If we take all elements from A and B: A has 1,2,3; B has 0,1,2. So combining all of them: 0,1,2,3. So that's four distinct elements. There's no 4 or anything else. So four elements. So answer is C:4. So yeah, I think that's the correct answer. The options are A:2, B:3, C:4, D:5. So the correct one is C.
</think> \\
<answer>\\
Given the sets $A=\{1,2,3\}$ and $B=\{0, 1, 2\}$, we need to find the number of elements in $A \cup B$.
First, we recall that the union of two sets $A$ and $B$ includes all elements that are in $A$, in $B$, or in both. Listing the elements of both sets and combining them while removing duplicates, we get: 
- Elements of $A$: 1, 2, 3 
- Elements of $B$: 0, 1, 2 
Combining these elements and removing duplicates, the union $A \cup B$ is $\{0, 1, 2, 3\}$. To verify, we use the formula for the number of elements in the union of two sets:
\[ |A \cup B| = |A| + |B| - |A \cap B| \]
where $|A| = 3$, $|B| = 3$, and $|A \cap B| = 2$ (since the intersection $A \cap B$ is $\{1, 2\}$). Plugging in these values:
\[ |A \cup B| = 3 + 3 - 2 = 4 \]
Thus, the number of elements in $A \cup B$ is 4, which corresponds to option C. \\
$\boxed{C}$ </answer> \\
\bottomrule
\end{tabularx}
\caption{\textbf{An example of a R1-style prompt-trajectory pair from SFT889K.} The expert trajectory features a reflective thinking process within \texttt{<think>} and \texttt{</think>} tags, followed by a structured solution within \texttt{<answer>} and \texttt{</answer>} tags. The final outcome is boxed.}
\label{tab:sft_data_exp}
\end{table*}

\newpage 
\begin{table*}[t]
\centering
\small
\begin{tabularx}{\linewidth}{>{\RaggedRight\arraybackslash}X}
\toprule
You are an expert mathematics evaluator. Your task is to analyze the reasoning complexity of a solution trajectory by breaking it down into distinct logical steps.\\
\textbf{Problem:}\\
\texttt{\{question\}}\\
\textbf{Thinking Trajectory:}\\
\texttt{\{think\}}\\
\textbf{Task:}\\
1. Analyze the trajectory above. Break down the reasoning process into a numbered list of distinct logical steps.\\
\quad $\bullet$ A \textbf{valid step} includes: a mathematical deduction, a calculation, a logical transition, setting up an equation, or verifying a condition.\\
\quad $\bullet$ \textbf{Exclude}: Simple restatements of the question, pure filler phrases (e.g., ``Let me see'', ``Hmm''), or redundant self-corrections that do not advance the logic.\\
2. Count the total number of valid steps derived in your list.\\
3. Output the final count inside a box format.\\
\textbf{Output Format:}\\
\texttt{[Your step-by-step analysis here...]}\\
\texttt{Total Logical Steps:} $\boxed{\text{number}}$\\
\bottomrule
\end{tabularx}
\caption{\textbf{Instruction prompt for logical-step segmentation.} Placeholders \texttt{\{question\}} and \texttt{\{think\}} are filled at inference time with the problem text and the model's thinking trajectory, respectively.}
\label{tab:logical_steps_prompt}
\end{table*}

\end{document}